\documentclass{article}

\usepackage[preprint]{neurips_2024}

\usepackage[utf8]{inputenc} 
\usepackage[T1]{fontenc}    
\usepackage{hyperref}       
\usepackage{url}            
\usepackage{booktabs}       
\usepackage{amsfonts}       
\usepackage{nicefrac}       
\usepackage{microtype}      
\usepackage{amsmath}
\usepackage{amsthm}
\usepackage{graphicx}
\usepackage{pifont}
\usepackage{bbm}
\usepackage{bm}
\usepackage{caption}
\usepackage[dvipsnames]{xcolor}

\usepackage{amsmath}
\usepackage{amssymb}
\usepackage{mathtools}
\usepackage{amsthm}
\usepackage{amssymb}

\newcommand{\cmark}{\ding{51}}%
\newcommand{\xmark}{\ding{55}}%

\newtheorem{theorem}{Theorem}[section]
\newtheorem{proposition}[theorem]{Proposition}
\newtheorem{lemma}[theorem]{Lemma}

\theoremstyle{definition}

\newtheorem{assumption}[theorem]{Assumption}
\usepackage{multirow}
\usepackage{adjustbox}
\usepackage[linesnumbered,ruled]{algorithm2e}

\title{Consistency Model is an Effective Posterior Sample Approximation for Diffusion Inverse Solvers}

\author{%
  Tongda Xu, Ziran Zhu, Jian Li, Dailan He, Yuanyuan Wang, Ming Sun\\
  Tsinghua University, Sensetime Research\\
  \texttt{x.tongda@nyu.edu} \\
  \And
  Ling Li, Hongwei Qin, Yan Wang, Jingjing Liu, Ya-Qin Zhang \\
  Kuaishou Technology, Chinese Academy of Sciences\\
  \texttt{wangyan@air.tsinghua.edu.cn} \\
}

\begin{document}

\maketitle

\begin{abstract}
Diffusion Inverse Solvers (DIS) are designed to sample from the conditional distribution $p_{\theta}(X_0|y)$, with a predefined diffusion model $p_{\theta}(X_0)$, an operator $f(\cdot)$, and a measurement $y=f(x'_0)$ derived from an unknown image $x'_0$. Existing DIS estimate the conditional score function by evaluating $f(\cdot)$ with an approximated posterior sample drawn from $p_{\theta}(X_0|X_t)$. However, most prior approximations rely on the posterior means, which may not lie in the support of the image distribution, thereby potentially diverge from the appearance of genuine images. Such out-of-support samples may significantly degrade the performance of the operator $f(\cdot)$, particularly when it is a neural network. In this paper, we introduces a novel approach for posterior approximation that guarantees to generate valid samples within the support of the image distribution, and also enhances the compatibility with neural network-based operators $f(\cdot)$. We first demonstrate that the solution of the Probability Flow Ordinary Differential Equation (PF-ODE) with an initial value $x_t$ yields an effective posterior sample $p_{\theta}(X_0|X_t=x_t)$. Based on this observation, we adopt the Consistency Model (CM), which is distilled from PF-ODE, for posterior sampling. Furthermore, we design a novel family of DIS using only CM. Through extensive experiments, we show that our proposed method for posterior sample approximation substantially enhance the effectiveness of DIS for neural network operators $f(\cdot)$ (e.g., in semantic segmentation). Additionally, our experiments demonstrate the effectiveness of the new CM-based inversion techniques. The source code is provided in the supplementary material.

\end{abstract}
\vspace{-1em}
\begin{figure}[thb]
\centering
    \includegraphics[width=\linewidth]{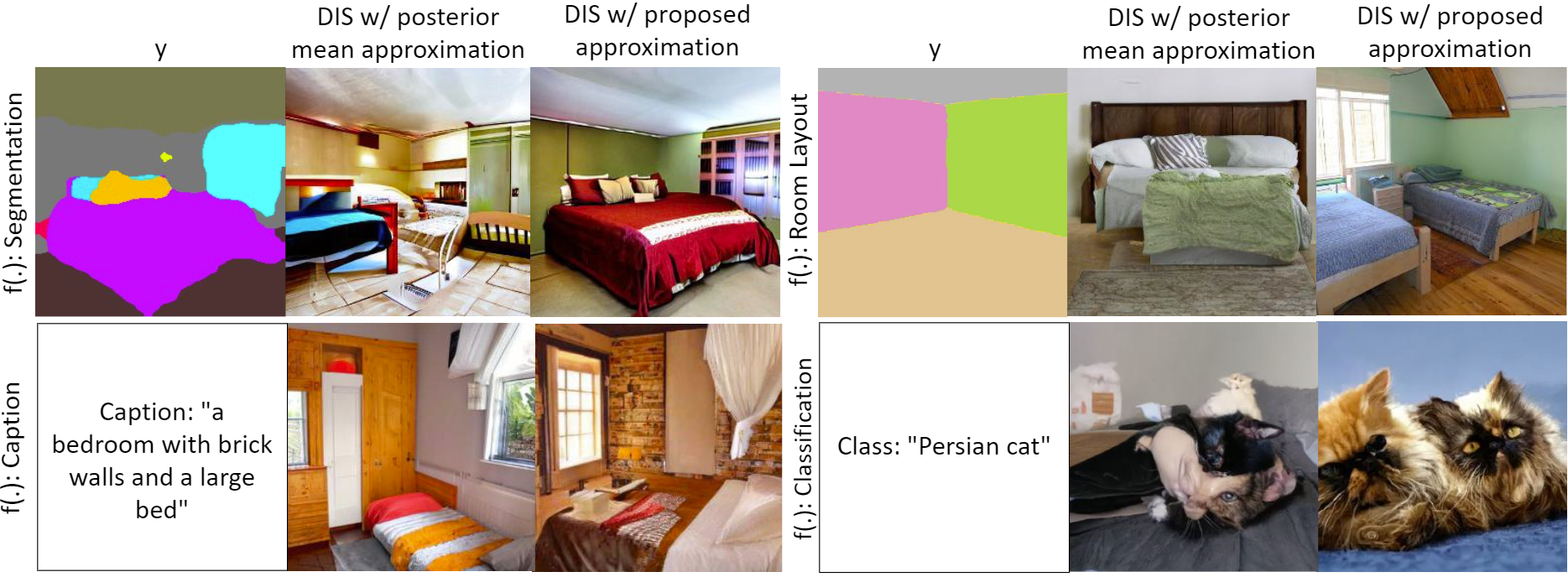}
\caption{A visual comparison of DIS with posterior mean as approximation for posterior sample, and DIS with proposed CM approximation for posterior sample.}
\label{fig:front}
\end{figure}
\vspace{-1em}
\section{Introduction}
Diffusion inverse solvers (DIS) are a family of algorithms that solve the inverse problem using diffusion prior \citep{li2023diffusion,moser2024diffusion}. More specifically, given an operator $f(.)$, a measurement $y = f(x'_0)$ from some unknown image $x'_0$, and a diffusion model $p_{\theta}(X_0)$, DIS aim to sample from the conditional distribution $x_0\sim p_{\theta}(X_0|y)$. For example, when $f(.)$ is a down-sampling operator, DIS is a perceptual super-resolution algorithm \citep{Menon2020PULSESP}. However, we cannot sample from  $p_{\theta}(X_0|y)$ directly as it is intractable. To tackle this challenge, previous works adopt a variety of techniques such as linear projection \citep{Wang2022ZeroShotIR,Kawar2022DenoisingDR,Chung2022ImprovingDM,Lugmayr2022RePaintIU,song2022pseudoinverse,pokle2024trainingfree,cardoso2024monte}, variational inference \citep{feng2023score,Mardani2023AVP,janati2024divide}, Bayesian filter \citep{dou2023diffusion}, sequential Monte Carlo \citep{wu2024practical,Phillips2024ParticleDD}, proximal gradient \citep{xu2024provably} and conditional score estimation \citep{Chung2022DiffusionPS,Yu2023FreeDoMTE,Zhu_2023_CVPR,He2023ManifoldPG,Song2023LossGuidedDM,boys2023tweedie,Rout2023BeyondFT,rout2024solving} for approximate or exact sampling.

Among those DIS techniques, the conditional score estimation methods \citep{Chung2022DiffusionPS, Song2023ConsistencyM} are most widely adopted, as they are suitable for general non-linear, noisy operator $f(.)$ and quite efficient in practice. During the inverse diffusion process from $X_T$ to $X_0$, they estimate the conditional score $\nabla_x\log p_{\theta}(X_t|y)$ by evaluating operator $f(.)$ with posterior sample from $p_{\theta}(X_0|X_t)$. As the posterior $p_{\theta}(X_0|X_t)$ is generally intractable, various approximations to posterior sample are proposed based on posterior mean: they either directly adopt posterior mean \citep{Chung2022DiffusionPS,Yu2023FreeDoMTE,Zhu_2023_CVPR,He2023ManifoldPG} or construct an uni-modal distribution centered at posterior mean \citep{Song2023LossGuidedDM,boys2023tweedie,Rout2023BeyondFT,rout2024solving}.

However, those posterior-mean based approximate posterior samples are far from real images, as it is well-known that the mean of noisy images may not lie in the support of the image distribution\citep{ledig2017photo,blau2018perception}. Although those out-of-distribution approximations are shown to be successful for simple $f(.)$ such as down-sampling and motion blurring, they might fail for more complex $f(.)$, especially when $f(.)$ are neural networks such as segmentation or classification, which are sensitive to out-of-distribution inputs.

In this paper, we propose a novel approach to approximate posterior sample for DIS. Our approximations are guarantee to be valid images and generally perform well for neural network $f(.)$. More specifically, we first show that given initial condition $X_t = x_t$, the solution of probability flow ordinary differential equation (PF-ODE) \citep{song2020score} is a valid posterior sample of true posterior $p_{\theta}(X_0|X_t = x_t)$. Inspired by this, we propose to use consistency model (CM), a distillation of PF-ODE as a posterior sample approximation for DIS. Furthermore, we propose a new family of DIS by iteratively inverting CM in a generative adversarial network (GAN) inversion fashion \citep{Creswell2016InvertingTG,Menon2020PULSESP}. Empirically, by using CM as posterior sample approximation, we improve the DIS's performance when operators are neural networks, such as semantic segmentation and image captioning. Furthermore, 
our experiments demonstrate the CM inversion also performs well for both neural network and non-neural network operators.

\section{Preliminaries}
\subsection{Diffusion Model} Diffusion model is a type of generative model with $T$ steps Gaussian Markovian chain in continuous space \citep{sohl2015deep}. Two widely adopted diffusion models are variance preserving (VP) and variance exploding (VE) diffusion. We follow the formulation of VE diffusion \citep{song2020score}, and refer VP diffusion to \citet{ho2020denoising,Kingma2021VariationalDM}. We denote the source image as $X_0$, and the forward process of VE diffusion is a Markov chain:
\begin{gather}
    q(X_T,...,X_1|X_0) = \prod_{t=1}^T q(X_t|X_{t-1}) \textrm{, where } q(X_t|X_{t-1}) = 
\mathcal{N}(X_{t-1}, (\sigma_t^2 - \sigma_{t-1}^2) I),\label{eq:dfw}
\end{gather}
where $\sigma^2_t$ are hyper-parameters (typically called variance schedule). The reverse diffusion process is also a Markov chain, with transition kernel $p(X_{t-1}|X_t)$ depending on the score function $\nabla_{X_t}\log p(X_t)$. To learn such a diffusion model, one can use a neural network $s_{\theta}(t,X_t)$ 
(parametrized by $\theta$)
to match the score function $\nabla_{X_t}\log p(X_t)$, and the resulting reverse diffusion process is given by
\begin{gather}
    p_{\theta}(X_0,...,X_T) = p(X_T)\prod_{t=1}^T p_{\theta}(X_{t-1}|X_t),\notag \\  \textrm{where } p_{\theta}(X_{t-1}|X_t) = \mathcal{N}(X_t + (\sigma_t^2 - \sigma_{t-1}^2)s_{\theta}(t,X_t),(\sigma_t^2 - \sigma_{t-1}^2)I). 
\end{gather}
\citet{song2020score} show that the reverse diffusion can be seen as a discretization of reverse stochastic differential equation (SDE) \citep{anderson1982reverse}. Further, there exists a probability flow ordinary differential equation (PF-ODE) that has the same marginal distribution $p_{\theta}(X_t)$ as the reverse SDE:
\begin{gather}
    \textrm{reverse SDE: } dX_t = -\frac{d \sigma^2_t}{d_t} s_{\theta}(t,X_t) dt + \sqrt{\frac{d \sigma^2_t}{d_t}}dB_t  \quad \overset{\textrm{same } p_{\theta}(X_t)}
    {\Longleftrightarrow} \notag \\ \textrm{PF-ODE: } dX_t = -\frac{1}{2}\frac{d\sigma^2_t}{dt}s_{\theta}(t,X_t)dt,
    \label{eq:rsde}
\end{gather}
where $B_t$ is the standard Brownian motion. Therefore, solving either of them is equivalent to sampling from the reverse diffusion process. Another useful result is Tweedie's formula \citep{Efron2011TweediesFA}, which provides an efficient estimation to the mean of posterior $p_{\theta}(X_0|X_t)$:
\begin{gather}
    \mathbb{E}[X_0|X_t] = X_t + \sigma^2_t s_{\theta}(t,X_t).\label{eq:twd}
\end{gather}
\subsection{Diffusion Inverse Solvers with Conditional Score Estimation} Given an operator $f(.)$, a target measurement $y = f(x'_0)$ from an unknown $x'_0$ and a diffusion model $p_{\theta}(X_0)$, the diffusion inverse solvers (DIS) attempt to sample from the conditional distribution $p_{\theta}(X_0|y)$. In this paper, we focus on DIS with conditional score estimation \citep{Chung2022DiffusionPS,Song2023LossGuidedDM}. More specifically, this paradigm of DIS attempts to estimate the conditional score $\nabla_{X_t}\log p_{\theta}(X_t|y)$. With this conditional score at hand, sampling from $p_{\theta}(X_0|y)$ is as easy as solving the reverse SDE or PF-ODE in Eq.~\ref{eq:rsde} with $s_{\theta}(t,X_t)$ replaced by the conditional score.

More specifically, \citet{Chung2022DiffusionPS,Song2023ConsistencyM} propose to expand the conditional score into unconditional score and a term that is related to a distance $d(f(x_{0|t}), y)$, where $x_{0|t}$ is the sample from posterior $p_{\theta}(X_0|X_t)$ and $d(.,.)$ is a distance:
\begin{gather}
    \nabla_{X_t}\log p_{\theta}(X_t|y) = \nabla_{X_t}\log p_{\theta}(y|X_t) + \nabla_{X_t}\log p_{\theta}(X_t), \notag \\
    \nabla_{X_t}\log p_{\theta}(y|X_t) = \nabla_{X_t}\log \mathbb{E}_{p_{\theta}(X_0|X_t)}[p_{\theta}(y|X_0)]
    \approx \nabla_{X_t}\log \sum_{x_{0|t}^{(i)}\sim p_{\theta}(X_0|X_t)}^{i=1,...,K} p_{\theta}(y|X_0=x_{0|t}^{(i)}), \notag \\
    p_{\theta}(y|X_0=x_{0|t}^{(i)}) \propto \exp{-d(f(x_{0|t}^{(i)}), y)}\label{eq:dps}.
\end{gather}
Under this formulation, an important issue is how to effectively draw differentiable samples from posterior $p_{\theta}(X_0|X_t)$. Obviously, direct ancestral sampling from reverse diffusion is computationally expensive. \citet{Chung2022DiffusionPS} propose to use the posterior mean computed by Tweedie's formula in Eq.~\ref{eq:twd} as
the posterior sample. \citet{song2022pseudoinverse, Song2023LossGuidedDM} propose to model the posterior as a Gaussian distribution with mean being the posterior mean as mean and and the covariance chosen as a hyper-parameter. \citet{Rout2023BeyondFT,boys2023tweedie} improve \citet{song2022pseudoinverse, Song2023LossGuidedDM} by using the posterior covariance computed by second order Tweedie's formula as the covariance of Gaussian. There are several other approaches that follow conditional score estimation paradigm \citep{Yu2023FreeDoMTE,Chung2023PrompttuningLD,song2023solving,He2023IterativeRB,rout2024solving,Meng2022DiffusionMB,dou2023diffusion,Chung2022ImprovingDM,song2022pseudoinverse,He2023FastAS} and rely on those approximations.

\section{Consistency Model is an Effective Posterior Sample Approximation for DIS}

\subsection{Previous Approximations are Out-of-Distribution} Most previous approximations to $p_{\theta}(X_0|X_t)$ either directly use the posterior mean or construct a uni-modal distribution entered around the posterior mean. However, the posterior mean $\mathbb{E}[X_0|X_t]$ is a mean-square error (MSE) minimizer for images perturbed by Gaussian noise, which does not necessarily correspond to a valid image \citep{ledig2017photo,blau2018perception}. In other words, the posterior mean may not lie in the support of natural image distribution, leading to its density in both marginal and posterior distributions approaching zero.:
\begin{gather}
    p_{\theta}(X_0=\mathbb{E}[X_0|X_t=x_t]) \approx 0, p_{\theta}(X_0=\mathbb{E}[X_0|X_t=x_t]|X_t=x_t) \approx 0
\end{gather}
\begin{figure}[thb]
\centering
\includegraphics[width=\linewidth]{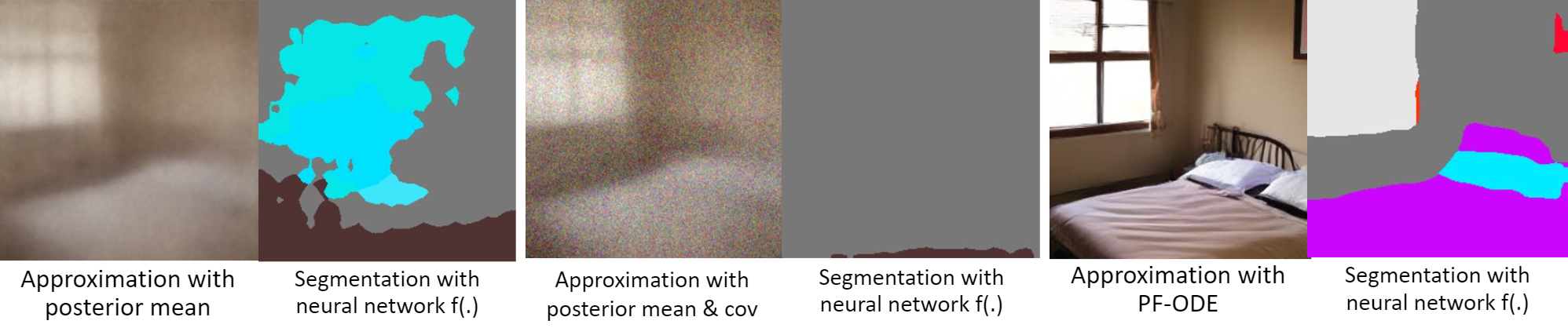}
\caption{Different approximations of posterior sample, and their output after a segmentation $f(.)$.}
\label{fig:psam}
\end{figure}
When the operator 
 is a neural network, such out-of-distribution approximations can
 significantly degrade the sample quality. A visual example is shown in Fig.~\ref{fig:psam}. The approximation using posterior mean, and the approximation using Gaussian centered at the posterior mean are not valid image samples. Consequently, when these approximations are processed through a semantic segmentation operator $f(.)$,
the outputs are nonsensical.

\subsection{PF-ODE provides an Effective Posterior Sample Approximation} It is known that the PF-ODE and reverse SDE in Eq.~\ref{eq:rsde} have the same marginal distribution, which is the image distribution $p_{\theta}(X_0)$ if the score is learned perfectly \citep{song2020score}. Denote the solution of PF-ODE given initial condition $X_t=x_t$ as $\Phi_0(x_t)$, then this solution is in support of natural image distribution, and the density of the solution is non-zero:
\begin{gather}
    p_{\theta}(X_0=\Phi_0(x_t)) > 0.
\end{gather}
Back to the previous example in Fig.~\ref{fig:psam}, when using PF-ODE as posterior sample approximation, the semantic segmentation operator $f(.)$ produces reasonable result. However, knowing $p_{\theta}(X_0=\Phi_0(x_t))>0$ is not enough. As we are seeking a posterior sample approximation, we need to ensure the solution $\Phi_0(x_t)$ has positive density on the posterior distribution, \textit{i.e.}, $p_{\theta}(X_0=\Phi_0(x_t)|X_t = x_t)>0$.

To the best of our knowledge, the relationship between PF-ODE's solution $\Phi_0(x_t)$ given initial value $X_t=x_t$, and the posterior $p_{\theta}(X_0|X_t=x_t)$ is not well understood. In this section, we show that the solution of PF-ODE has non-zero density in true posterior for any time $t$, \textit{i.e.,} 
\begin{gather}
    \forall x_t, p_{\theta}(X_0=\Phi_0(x_t)|X_t = x_t) > 0.
\end{gather}

\begin{assumption} 
\label{asm:01}
We assume the following conditions hold:
\begin{itemize}
    \item The distribution $p_{\theta}(X_0)$ can be approximated by a $d$-dimension Gaussian Mixture Model (GMM) composed of $N$ Gaussians with same small diagonal covariance $\sigma^2I$ and mean $\mu^i$:
\begin{gather}
    p_{\theta}(X_0) \approx \frac{1}{N}\sum_{i=1}^N\mathcal{N}(X_0|\mu^i,\sigma^2I).
\end{gather}
\item The solution $\Phi_0(x_t)$ and initial value $x_t$ are bounded, \textit{i.e.}, $||\Phi_0(x_t)||<c, ||x_t||\le c$.
\item As PF-ODE is margin preserving and the dimension of data is high, the solution of PF-ODE is close to the surface of a sphere centered at $\mu^i$ with radius $\sqrt{d}\sigma$ \citep{vershynin2018high}, \textit{i.e.}, $\exists k,  \textrm{s.t.} ||\Phi_0(x_t) - \mu^k||^2\le \sigma^2 + d\sigma^2$ with probability $1-p$ for small $p$.
\end{itemize}
\end{assumption}
\begin{proposition}
The solution of PF-ODE has a positive likelihood in true posterior with high probability, \textit{i.e.},
\begin{gather}
    p_{\theta}(X_0=\Phi(x_t)|X_t=x_t) \ge \frac{1}{N}\frac{1}{\sqrt{(4\pi\sigma^2)^d}}\exp{(-\frac{2c^2}{\sigma_t^2} - \frac{d+1}{2})} \textrm{, with probability } 1-p.
\end{gather}
\end{proposition}
Despite the solution of PF-ODE has a non-zero density in true posterior with high probability, we can not tell whether the solution falls into the highest density mode of true posterior. However, when $\sigma_t^2$ is small, we can show that the solution of PF-ODE is approximately the highest density mode of true posterior.
\begin{lemma}
    The PF-ODE can be written as:
    \begin{gather}
    \frac{dX_t}{dt} = \underbrace{\sum_{i=1}^N\frac{w_i}{2}\frac{d\sigma_t^2}{dt} \frac{(X_t-\mu^i)}{\sigma^2+\sigma_t^2}}_{\textrm{velocity field } v_t}, w_i= (\exp{-\frac{||X_t-\mu^i||^2}{2(\sigma^2+\sigma_t^2)}}) / (\sum_{j=1}^N\exp{-\frac{||X_t-\mu^j||^2}{2(\sigma^2+\sigma_t^2)}}) \label{eq:pfode}.
\end{gather}
\end{lemma}
\begin{minipage}[t]{0.5\linewidth}
\vspace{0pt}
\centering
\includegraphics[width=\linewidth]{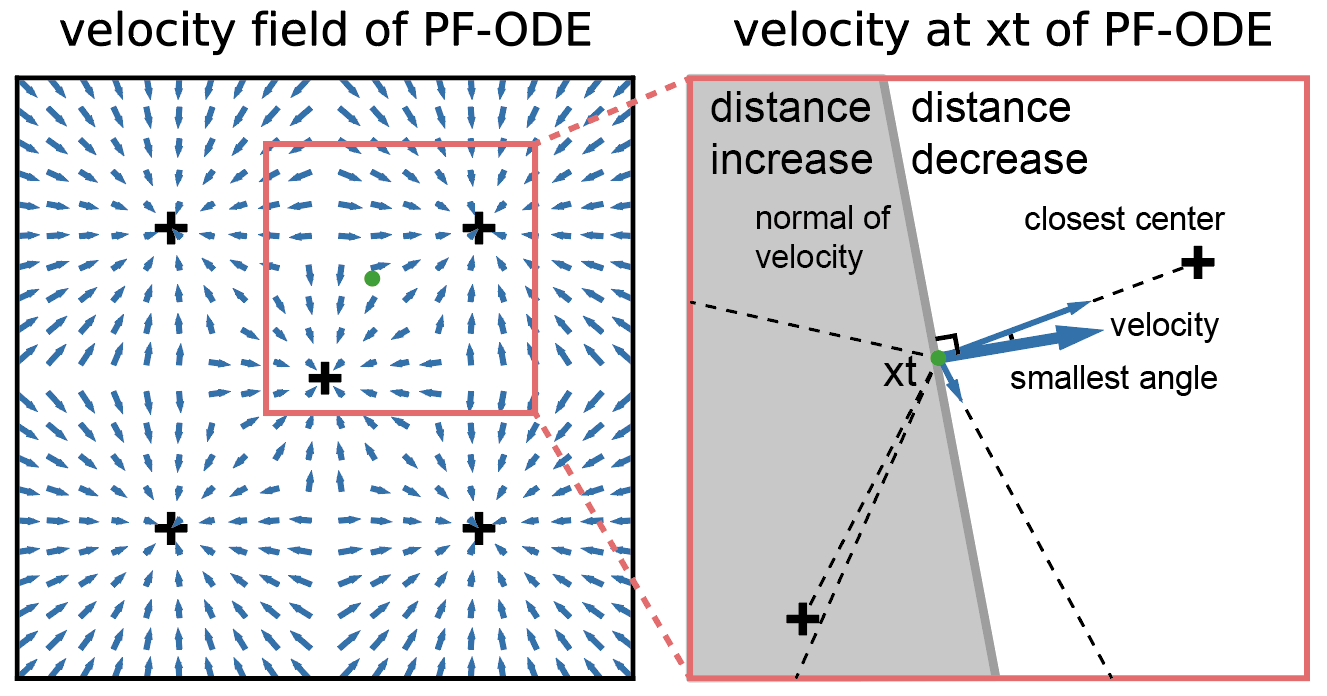}
\captionof{figure}{The PF-ODE's velocity field of a five GMM example.}
\label{fig:vf}
\end{minipage}
\begin{minipage}[t]{0.5\linewidth}
We can consider the velocity field $v_t$ of PF-ODE in Eq.~\ref{eq:pfode}. The velocity field is a sum of vectors pointing to centers $\mu^i$ weighted by soft-max function $w_i$. When $t$ is small, $w_i$ becomes a "hard"-max which selects the closest center $\mu^*$ to initial point $x_t$, and the velocity field always points to $\mu^*$. On the other hand, this closest center $\mu^*$ is also the highest density mode in true posterior (See Appendix.~\ref{app:pf}). With this coincidence, the solution of PF-ODE with $X_t=x_t$ is approximately the mode with highest density in true posterior.
\end{minipage}

When $\sigma_t^2$ is not small enough, we can still consider other conditions when PF-ODE will converge to the closest center $\mu^*$. Consider the five GMM example in Figure.~\ref{fig:vf} with initial point $X_t=x_t$. An obvious intuition is that the normal plane of velocity $v_t$ divide the space into two parts. In one part, the velocity has a negative projection on center's direction. In the other part, the velocity has a positive projection. Then, $X_t$ will move away from the centers with negative projection, to the centers with positive projection. Among the centers with positive projection, when the closest center $\mu^*$ also has the smallest angle with velocity, it is very likely that the PF-ODE eventually converges into the closest center.

\subsection{A Toy Example}
\begin{figure}[thb]
\centering
    \includegraphics[width=\linewidth]{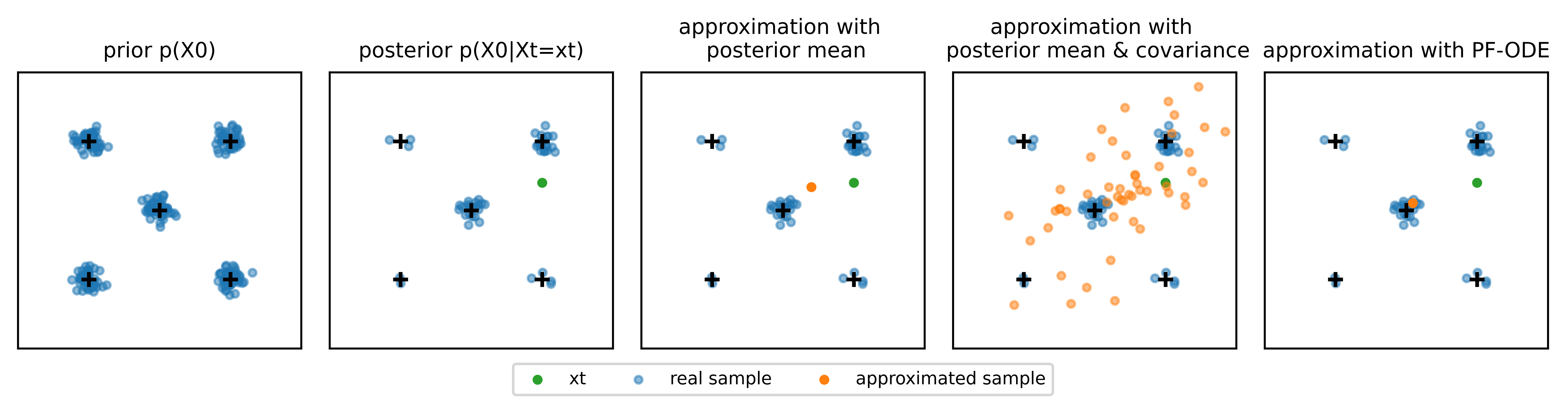}
    \includegraphics[width=\linewidth]{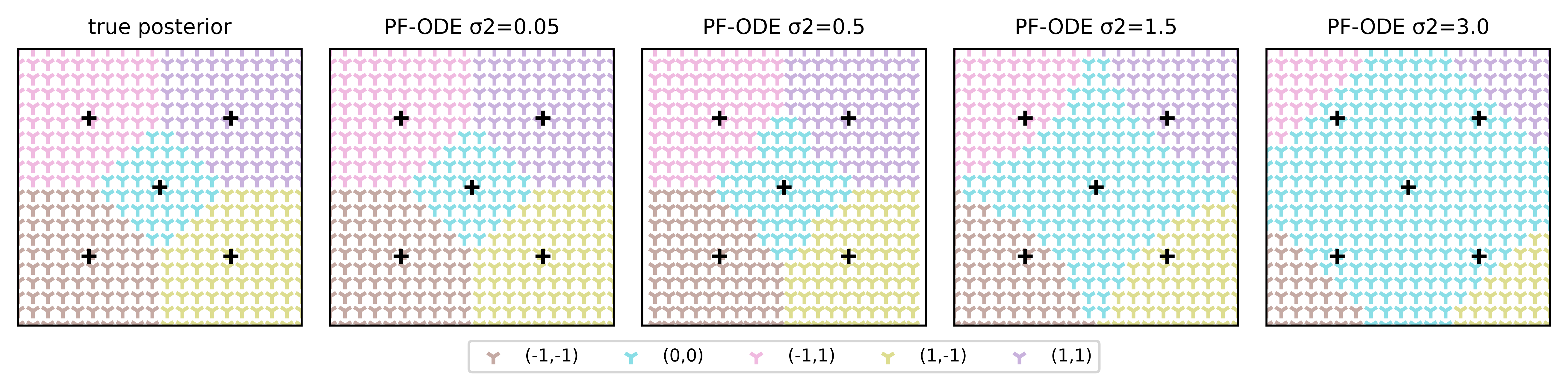}
\caption{A toy example with five GMM.}
\label{fig:toy}
\end{figure}

To better understand the results above, we provide a toy example in $\mathbb{R}^2$. As shown in Fig.~\ref{fig:toy}, the source distribution $p(X_0)$ is a five Gaussian mixture model (GMM). Each Gaussian is diagonal with standard deviation $\sigma_0=0.1$. The centers of Gaussians are $(-1,-1),(-1,1),(1,1),(1,-1),(0,0)$. We adopt VE diffusion with $\sigma_T=4$, $T=100$ and $\sigma$ schedule in \citet{Karras2022ElucidatingTD}. The score function $\nabla_{X_t}\log p(X_t)$ is computed analytically.

As shown in Fig.~\ref{fig:toy}.upper, starting from $x_t=(1,0.4)$, the posterior $p(X_0|X_t=x_t)$ has most density on two right-hand side Gaussians. However, the approximation with posterior mean is close to $(1,0.4)$, where true posterior has almost no density. Furthermore, the approximation with posterior mean and covariance also concentrates on a region where true posterior has almost no density. However, the approximation with PF-ODE falls into a high density region of true posterior. 

We also visualize the highest density mode's decision boundary of true posterior and PF-ODE starting at different $\sigma^2_t$. As shown in Fig.~\ref{fig:toy}.lower, the decision boundary of true posterior is always a Voronoi cell centered at $\mu^i$. When $\sigma_t^2$ is small, the solution of PF-ODE is similar to the true posterior. As $\sigma_t^2$ increase, the solution of PF-ODE becomes less similar to true posterior. However, at that time, the density scatters more evenly in the true posterior. The solution of PF-ODE still has non-zero density. 

\subsection{Implementation of PF-ODE Approximation with Consistency Model}
\label{sec:dpscm}
Directly solving PF-ODE is also intractable for DIS. Fortunately, PF-ODE can be distilled by Consistency Model (CM) \citep{Song2023ConsistencyM}. More specifically, CM trains a one-step neural function $g_{\theta}(t,x_t)$ to approximate the solution of PF-ODE $\Phi_0(x_t)$. Its gradient is cheap to evaluate. Therefore, we can directly replace the $x_{0|t}\sim p_{\theta}(X_0|X_t=x_t)$ step in Eq.~\ref{eq:dps} by $x_{0|t} = g_{\theta}(t,x_t)$.

In practice, we find CM often over-fits the operator $f(.)$. More specifically, the CM approximated sample $x_{0|t}$ is numerically close to $y$ after passing the operator $f(.)$. However, by inspecting $x_{0|t}$ visually, one often concludes that it is not aligned with $y$. (See an example in Fig.~\ref{fig:of}).

In fact, the resulting overfitted sample is an adversial example \citep{szegedy2013intriguing}, which is aligned with label $y$ according to neural network but not aligned with $y$ according to human eye. To make $f(.)$ robust, we propose to add a small Gaussian noise to the output of CM as literature in adverisal robustness \citep{li2019certified} 
\begin{gather}
    x_{0|t} = g_{\theta}(t,x_t) + \mathcal{N}(0,\tau^2).
\end{gather}


\subsection{Consistency Model Inversion}
\label{sec:cmi}
We have shown that CM can be used as posterior sample approximator for DIS. As CM can be seen as a conditional GAN conditioned on timestep $t$ and state $x_t$, a natural question to ask is: can CM be inverted in a GAN inversion fashion \citep{Creswell2016InvertingTG,Menon2020PULSESP}?

We first review GAN inversion briefly. GAN inversion optimizes the noise $z$ by penalizing the distance between generated image and the target for $K$ steps. And step by step, the optimized $z$ will generate an image that satisfies the constraint. Denote $h_{\theta}(.)$ as GAN, we have
\begin{gather}
    z^0\sim\mathcal{N}(0,I), x^i = h_{\theta}(z^i), z^{i+1} = z^i -\zeta \frac{d}{d z^i} d(f(x^i),y),i=0,...,K-1.
\end{gather}
Recall that CM first initializes an $x_T$ from Gaussian distribution, then transforms it into target image $x_{0|t}$ by a neural network. Then iteratively CM adds a small noise back to $x_{0|t}$ and denoises again. For each iteration, CM is the same as GAN. To CM, for each iteration, we can invert CM as GAN inversion. In other words, we can stack several GAN inversions together, according to the iterative CM sampling algorithm in Algorithm~\ref{alg:cm}, to obtain our final CM inversion algorithm in Algorithm~\ref{alg:cmi}. For each iteration, a GAN inversion is performed. Similarly, we also find that adding a small Gaussian noise is beneficial to neural-network $f(.)$.

\section{Experiments}
\label{sec:exp}
\subsection{Experiment Setup}
\textbf{Base Diffusion Models}
For diffusion model, we use a pretrained VE diffusion - EDM \citep{Karras2022ElucidatingTD} provided by \citet{Song2023ConsistencyM}. For EDM-related methods, we adopt ancestral sampler with 1000 Euler steps. For CM \citep{Song2023ConsistencyM}, we employ the official pre-trained model by \cite{Song2023ConsistencyM}. The details are shown in Appendix.~\ref{app:im}.

\textbf{Operators} We evaluate all the methods with four neural network operators: semantic segmentation, room layout estimation, image captioning and image classification. For layout estimation, we adopt the neural network by \citet{Lin2018IndoorSL}. For semantic segmentation, we use the neural network by \citet{zhou2017scene}. For image captioning, we employ BLIP \citep{li2022blip}. For image classification, we use ResNet \citep{He2015DeepRL}. In addition, we also evaluate a simple non-neural network operator: down-sampling (x4).

\textbf{Datasets \& Metrics}
Following \citet{Song2023ConsistencyM} and \citet{Chung2022DiffusionPS}, we use the first 1000 image from LSUN Bedroom and LSUN Cat dataset \citep{yu15lsun} as test set. All images are resized into $256^2$. To evaluate sample quality, we use Fréchet Inception Distance \citep{Heusel2017GANsTB} and Kernel Inception Distance (KID) \citep{Binkowski2018DemystifyingMG}. To evaluate consistency with the constraint, we use mIOU for segmentation and layout, CLIP score for captioning, and Accuracy for classification. For neural network $f(.)$, we use different models for DIS and testing (See Appendix.~\ref{app:im}). For down-sampling, we use image restoration metrics such as LPIPS \citep{Zhang2018TheUE} and peak signal-to-noise ratio (PSNR).

\textbf{Previous State-of-the-Art DIS}
We compare our approach with previously published DIS that are able to solve neural network operator $f(.)$. For methods that directly use posterior mean as posterior sample, we include DPS \citep{Chung2022ImprovingDM}, FreeDOM \cite{Yu2023FreeDoMTE} and MPGD \citep{he2023manifold}. For methods that construct an approximated posterior distribution with posterior mean as mode, we include LGD \citep{Song2023LossGuidedDM} and STSL \citep{Rout2023BeyondFT} (See Tab.~\ref{tab:pa}). We implement all those methods with EDM and Euler ancestral sampler (See details in Appendix.~\ref{app:im}). We acknowledge that there are other very competitive works designed for latent diffusion  \citep{Chung2023PrompttuningLD,song2023solving,He2023IterativeRB,rout2024solving} or linear operator $f(.)$ \citep{Meng2022DiffusionMB,dou2023diffusion,Chung2022ImprovingDM,song2022pseudoinverse,boys2023tweedie,He2023FastAS}. However, for now we focus on pixel domain diffusion with neural network $f(.)$, and have not included them for comparison.

\begin{table}[t]
\begin{minipage}[t]{\linewidth}
\vspace{0pt}
\centering
\caption{Results on neural network operators, \textit{i.e.}, layout estimation, segmentation, caption and classification. \textbf{Bold}: best in diffusion-based DIS. \underline{Underline}: second best in diffusion-based DIS.}
\label{tab:high}
\resizebox{\linewidth}{!}{
\begin{tabular}{@{}lcccccccccccc@{}}
\toprule
& \multicolumn{9}{c}{LSUN Bedroom} & \multicolumn{3}{c}{LSUN Cat} \\ \cmidrule(ll){2-10} \cmidrule(ll){11-13}
            & \multicolumn{3}{c}{Segmentation} & \multicolumn{3}{c}{Layout} & \multicolumn{3}{c}{Caption} & \multicolumn{3}{c}{Classification} \\ \cmidrule(ll){2-4}\cmidrule(ll){5-7} \cmidrule(ll){8-10} \cmidrule(ll){11-13}
            & mIOU        & FID        & KID       & mIOU       & FID       & KID      & CLIP       & FID        & KID        & Acc       & FID      & KID      \\ \midrule
\multicolumn{3}{@{}l@{}}{\textit{Diffusion Model based}} &           &                   &           &          &                 &            &            &                  &          &          \\
EDM (Base)  & 0.17 & 6.35 & 1.4e-3 & 0.38 & 6.35 & 1.4e-3 & 21.27 & 6.35 & 1.4e-3 & 0.14 & 10.26 & 3.1e-3 \\
DPS & 0.27 & 22.84 & 1.0e-2 & 0.54 & \underline{7.59} & \textbf{1.9e-3} & 22.57 & 9.49 & 2.6-e3 & 0.79 & 15.73 & 7.1e-3 \\ 
FreeDOM & \underline{0.27} & 21.90 & 9.1-e3 & 0.46 & 15.27 & 8.1e-3 & \underline{22.61} & 28.30 & 1.8e-2  & \underline{0.84} & 32.32 & 1.7-e2 \\
MPGD     & 0.24 & 82.66 & 6.9e-2 & \underline{0.73} & 15.38 & 8.5-e3& 21.49 & 21.14 & 1.3e-2 & 0.37 & 15.40 & 7.1e-3 \\
LGD & 0.22 & 35.69 & 2.4e-2 & 0.70 & 8.07 & 2.3e-3 & 22.58 & \underline{8.38} & \underline{2.6-e3} & 0.64 & \textbf{13.35} & \underline{4.5e-3} \\
STSL & \underline{0.27} & \underline{19.48} & \textbf{7.4e-3}  & 0.52 & 7.74 & \underline{2.2e-3} & 22.39 & 9.70 & 2.8e-3 & 0.78 & 15.74 & 6.9e-3 \\
Proposed I  & \textbf{0.34} & \textbf{18.06} & \underline{8.2e-3} & \textbf{0.78} & \textbf{7.50} & \underline{2.2e-3}  & \textbf{22.63} & \textbf{8.16} & \textbf{2.5-e3} & \textbf{0.90} & \underline{13.45} & \textbf{3.6e-3} \\ \midrule
\multicolumn{3}{@{}l@{}}{\textit{Consistency Model based}} & &                   &           &          &                 &            &            &            &          &          \\
CM (Base)  & 0.18 & 20.45 & 1.0e-2 & 0.37 & 20.45 & 1.0e-2  & 21.40 & 20.45 & 1.0e-2 & 0.12 & 27.15 & 1.3e-2 \\
Proposed II & 0.32 & 32.60 & 2.2e-2 & 0.82 & 15.43 & 8.1e-3 & 22.56 & 14.86 & 6.7e-3 & 0.92 & 27.35 & 1.4e-2 \\ \bottomrule
\end{tabular}
}
\end{minipage}
\begin{minipage}[t]{0.62\linewidth}
\vspace{0pt}
\centering
\caption{The posterior sample approximation of different methods.}
\label{tab:pa}
\resizebox{\linewidth}{!}{
\begin{tabular}{@{}lcc@{}}
\toprule
                     & Approximation of posterior sample  
& Valid image?\\ \midrule
DPS & $x_{0|t} = \mathbb{E}[X_0|X_t]$ & \xmark \\ 
FreeDOM & $x_{0|t} = \mathbb{E}[X_0|X_t]$ & \xmark \\ 
MDPG & $x_{0|t} = \mathbb{E}[X_0|X_t]$ & \xmark \\ 
LGD & $x_{0|t} \sim \mathcal{N}(\mathbb{E} [X_0|X_t],r_t^2I) $ & \xmark \\ 
STSL & $x_{0|t}\sim \mathcal{N}(\mathbb{E} [X_0|X_t],\textrm{Cov}(X_0|X_t))$ & \xmark \\ 
Proposed I & $x_{0|t} = g_{\theta}(t,X_t)$ & \cmark
 \\ \bottomrule
\end{tabular}
}
\end{minipage}
\hfill
\begin{minipage}[t]{0.37\linewidth}
\centering
\vspace{0pt}
\captionof{table}{Temporal and spatial complexity of different methods.}
\label{tab:com}
\resizebox{\linewidth}{!}{
\begin{tabular}{@{}lcc@{}}
\toprule
            & Time (s) & VRAM (GB) \\ \midrule
DPS         & 150 & 5.35 \\
Proposed I  & 218 & 6.32 \\
Proposed II & 72 & 6.58 \\ \bottomrule
\end{tabular}
}
\end{minipage}\hfill
\end{table}
\subsection{Main Results}
\begin{figure*}[htb]
\centering
    \includegraphics[width=\linewidth]{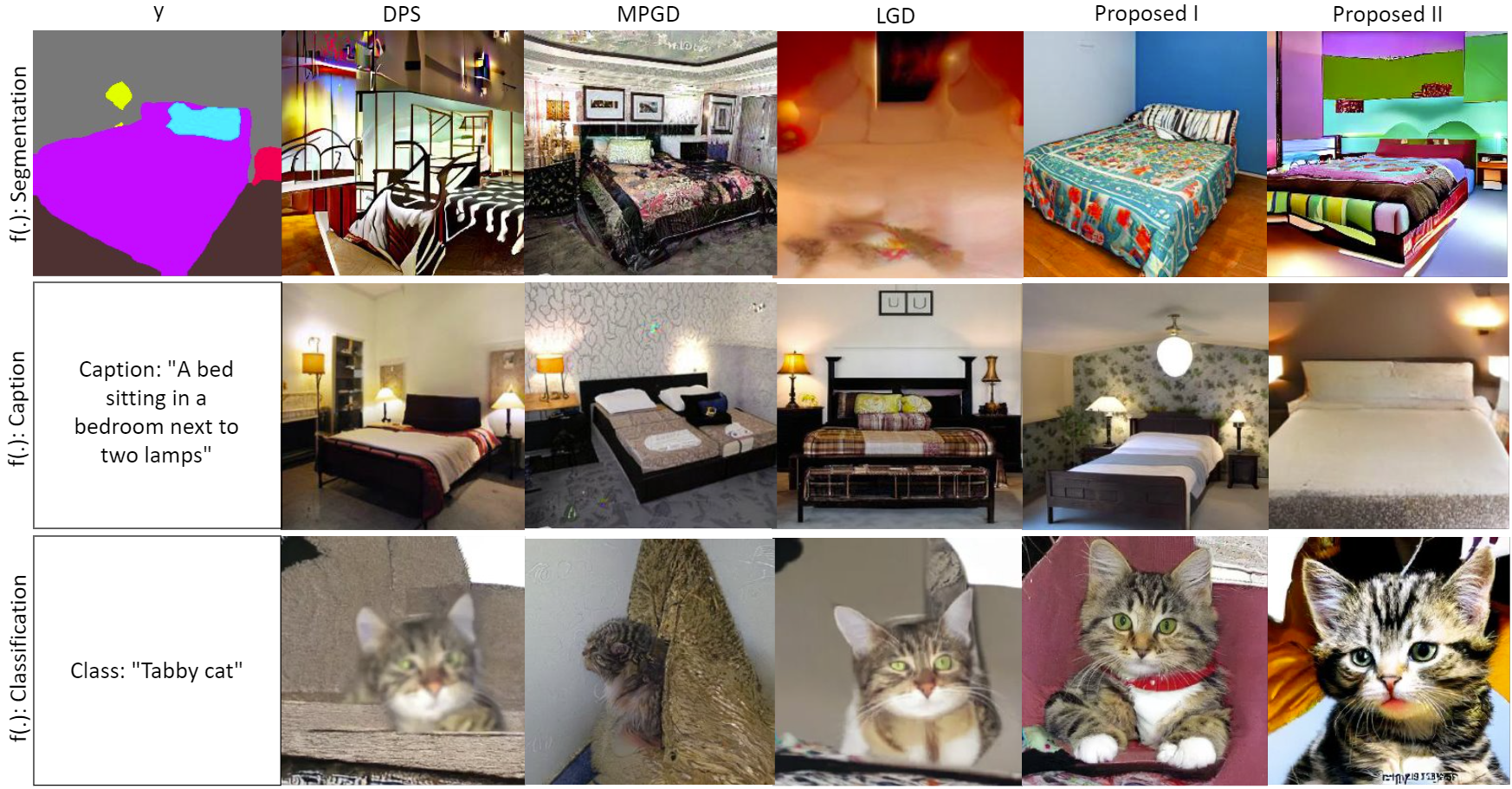}
\caption{Visual results on neural network operators such as segmentation, caption and classification.}
\label{fig:vr}
\end{figure*}

\textbf{Results on Neural Network Operators}
We test our proposed approaches on four neural network operators: segmentation, layout estimation, caption and classification. As shown in Tab.~\ref{tab:high}, Fig.~\ref{fig:front}, Fig.~\ref{fig:vr}, Fig.~\ref{fig:avr1} and Fig.~\ref{fig:avr3}, both quantitatively and visually, our Proposed I (Sec.~\ref{sec:dpscm}) has significant improvement on both consistency (\textit{e.g.}, mIOU) and sample quality (\textit{e.g.}, FID) over the baseline DPS \citep{Chung2022DiffusionPS}. The advantage of our approximation over other posterior mean based approximations is clearly demonstrated. This is because neural network $f(.)$ are sensitive to out-of-distribution input. On the other hand, our Proposed II (Sec.~\ref{sec:cmi}) is also quite effective compared with unconditional CM.

\textbf{Results on Non-neural Network Operators}
In additional to neural network operators, we also verify that our approaches work well for non-neural network operators such as down-sampling. Results are summarized in Tab.~\ref{tab:rlow} and Fig.~\ref{fig:of}.\textit{lower}. Our Proposed I is only comparable to DPS for simple operators. This is because non-neural network $f(.)$ are not that sensitive to out-of-distribution approximations. Besides, our Proposed II also works well for linear operator.

\subsection{Ablation Study}
We evaluate the effect of using CM for posterior approximation in Proposed I (Sec.~\ref{sec:dpscm}) and adding randomness to CM in Proposed I \& II (Sec.~\ref{sec:dpscm}, Sec.~\ref{sec:cmi}) in Tab.~\ref{tab:abl}. For Proposed I, we show that using CM to replace posterior mean reduces the distance with measurement $y$ in mIOU and improves the sample quality in FID. Similarly, for both Proposed I \& II, adding randomness improves mIOU and reduces FID.

\begin{minipage}[t]{\linewidth}
\begin{minipage}[t]{0.49\linewidth}
\vspace{0pt}
\centering
\captionof{table}{Ablation study of Proposed I and Proposed II with Bedroom segmentation. \textbf{Bold}: Method with best performance.}
\label{tab:abl}
\resizebox{0.85\linewidth}{!}{
\begin{tabular}{@{}lcccc@{}}
\toprule
                             & CM & Rand & mIOU & FID \\ \midrule
\multirow{3}{*}{Proposed I}  & \xmark & \xmark & 0.27 & 22.84 \\
                             & \cmark & \xmark & 0.31 & 19.29 \\
                             & \cmark & \cmark & \textbf{0.34} & \textbf{18.06} \\ \midrule
\multirow{2}{*}{Proposed II} & - & \xmark & 0.29 & 32.93    \\
                             & - & \cmark & \textbf{0.32} & \textbf{32.60} \\ \bottomrule
\end{tabular}
}
\end{minipage}
\hfill
\begin{minipage}[t]{0.49\linewidth}
\vspace{0pt}
\centering
\captionof{table}{Ablation study of randomness and data augmentation with segmentation.}
\label{tab:abl2}
\resizebox{0.9\linewidth}{!}{
\begin{tabular}{@{}lccc@{}}
\toprule
\multirow{2}{*}{Rand} & \multicolumn{3}{c}{mIOU}             \\ \cmidrule(l){2-4} 
                            & Model A & Model A + DA & Model B \\ \midrule
\xmark & 0.57 & 0.43 & 0.31 \\
\cmark & 0.51 & 0.54 & 0.34 \\ \bottomrule
\end{tabular}
}
\end{minipage}\hfill
\end{minipage}
\begin{table}[t]
\begin{minipage}[t]{0.54\linewidth}
\vspace{0pt}
\centering
\includegraphics[width=\linewidth]{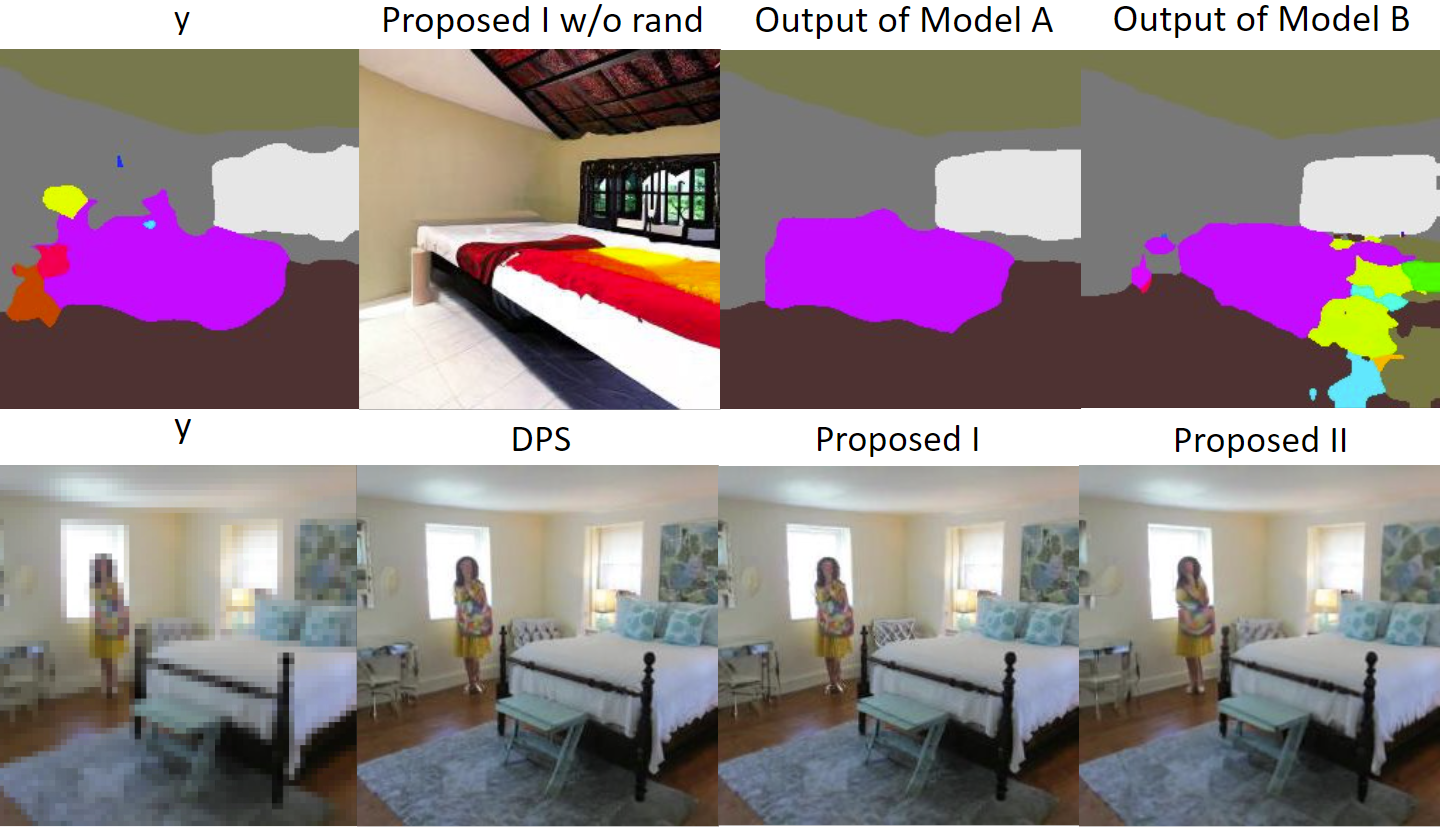}
\captionof{figure}{\textit{upper}. An example of over-fitting an operator $f(.)$. \textit{lower}. Visual results on linear operators such as down-sampling.}
\label{fig:of}
\end{minipage}
\hfill
\begin{minipage}[t]{0.45\linewidth}
\centering
\vspace{0pt}
\captionof{table}{Results on non-neural network operators such as super-resolution.}
\label{tab:rlow}
\resizebox{\linewidth}{!}{
\begin{tabular}{@{}lcccc@{}}
\toprule
            & \multicolumn{4}{c}{Bedroom Down-sampling (x4)} \\ \cmidrule(ll){2-5}
            & KID & FID & LPIPS & PSNR \\ \midrule
\multicolumn{3}{@{}l@{}}{\textit{Diffusion Model based}} &  \\
EDM (Base)  & 1.4e-3 & 6.35 & 0.73 & 9.10 \\
DPS & 1.4e-3 & \underline{4.82} & \textbf{0.11} & 26.69 \\
FreeDOM & \textbf{1.1e-3} & 4.84 & \textbf{0.11} & \underline{26.70} \\
MPGD & 2.0e-3 & 43.36 & 0.38 & 22.84 \\
LGD & 2.8e-3 & 7.10 & 0.14 & 26.51 \\
STSL & \underline{1.2e-3} & \textbf{4.80} & \underline{0.12} & 26.65 \\
Proposed I  & 1.9e-3 & 5.67 & \underline{0.12} & \textbf{26.91} \\\midrule
\multicolumn{3}{@{}l@{}}{\textit{Consistency Model based}} &  \\
CM (Base)  & 1.0e-2 &  20.45 & 0.73 & 9.49 \\
Proposed II & 2.9e-3 & 6.66 & 0.14 & 26.45 \\ \bottomrule
\end{tabular}
}
\end{minipage}\hfill
\end{table}

We assume that CM benefits from randomness as it avoids overfitting the operator $f(.)$, or it makes $f(.)$ robust to adversial examples \citep{li2019certified}. To verify this assumption, we use Model A for $f(.)$ during DIS. During testing $f(.)$, we compare the results of Model A, Model A w/ data augmentation (DA), and a separate Model B. In Tab.~\ref{tab:abl2}, we show that when tested with Model A, the DIS w/o randomness outperform  DIS w/ randomness. While on Model A w/ DA and Model B, the DIS w/ randomness outperforms DIS w/o randomness. This indicates that DIS w/o randomness overfits Model A. An example of such overfitting is presented in Fig.~\ref{fig:of}.\textit{upper}. 

\subsection{Complexity}
In Tab.~\ref{tab:com}, we show that Proposed I is $30\%$ slower than DPS while Proposed II is $50\%$ faster. Both of our approaches cost $1$ GB more VRAM. 

\section{Related Work}
\textbf{Diffusion Inverse Solvers} An important branch of DIS focus on linear operator $f(.)$ with projection or pseudo-inverse \citep{Wang2022ZeroShotIR,Kawar2022DenoisingDR,Chung2022ImprovingDM,Lugmayr2022RePaintIU,song2022pseudoinverse,dou2023diffusion,pokle2024trainingfree,cardoso2024monte}. For general, non-linear $f(.)$, various approaches are proposed such as Monte carlo \citep{wu2024practical,Phillips2024ParticleDD}, proximal gradient \citep{xu2024provably} and variational inference \citep{feng2023score,Mardani2023AVP,janati2024divide}. Among those paradigms, the conditional score estimation methods are mostly adopted as they are scalable to practically large images with reasonable run-time \citep{Chung2022DiffusionPS,Yu2023FreeDoMTE,Zhu_2023_CVPR,He2023ManifoldPG,Song2023LossGuidedDM,boys2023tweedie,Rout2023BeyondFT,rout2024solving}. Following this paradigm, we propose to approximate posterior sample with PF-ODE, which improves the results for neural network $f(.)$.

\textbf{GAN Inverse Solvers} Similar to diffusion inversion, given the measurement and operator, GAN inversion finds an image $x$ on the prior of GAN by minimizing the distance of generated image and a target \citep{Creswell2016InvertingTG}. It is widely adopted for image editing and image restoration \citep{Menon2020PULSESP, Daras2021IntermediateLO}. We propose to treat CM as several stacked GANs and invert CM as a series of GAN inversions.

\section{Discussion \& Conclusion}
One major limitation of our work is that all the experiments are conducted with $256^2$ images and diffusion models in pixel domain. Recent works already show the potential of CM on $512^2$ for latent diffusion models \citep{Luo2023LatentCM,Chung2023PrompttuningLD,song2023solving,He2023IterativeRB,rout2024solving}. 
It would be interesting to see how our approach works empirically for large images using latent diffusion models.

To conclude, we show that the solution of PF-ODE is an effective posterior sample. Built upon this, we propose to use CM as a high-quality approximation to posterior sample. Further, we propose a new family of DIS using only CM. Experimental results show that our proposed approaches perform well for neural network operators.

\bibliography{main}

\clearpage

\appendix

\section{Proof of Main Results}
\label{app:pf}
We first derive some basic properties of the GMM model. More specifically, we have
\begin{gather}
    p_{\theta}(X_0) = \frac{1}{N}\sum_{i=1}^N\mathcal{N}(X_0|\mu^i,\sigma^2I), \\
    p_{\theta}(X_t|X_0) = \mathcal{N}(X_t|X_0,\sigma_t^2I), \\
    p_{\theta}(X_t) = \frac{1}{N}\sum_{i=1}^N\mathcal{N}(X_t|\mu^i,(\sigma^2+\sigma_t^2)I),\\
    p_{\theta}(X_0|X_t) = \frac{p_{\theta}(X_t|X_0)p_{\theta}(X_0)}{p_{\theta}(X_t)} \notag \\
    = \sum_{i=1}^N\frac{\mathcal{N}(X_t|X_0)}{\sum_{j=1}^N\mathcal{N}(X_t|\mu^j,(\sigma^2+\sigma_t^2)I)}\mathcal{N}(X_0|\mu^i,\sigma^2I) \notag \\
    = \sum_{i=1}^N u_i \mathcal{N}(X_0|\mu^i,\sigma^2 I), \notag \\
    \textrm{where } u_i = (\exp -\frac{||X_0-X_t||^2}{2\sigma_t^2})/(\sum_{j=1}^N\frac{1}{\sqrt{(1+\sigma^2/\sigma_t^2)^d}}\exp{-\frac{||X_t-\mu^i||^2}{2(\sigma^2+\sigma_t^2)}}).
\end{gather}
For true posterior $p_{\theta}(X_0|X_t=x_t)$, we know that $X_0$ eventually converges to one of the $\mu^i$. Therefore, the $\mu^*$ closest to $X_t$ will have highest weighting $u_i$. And thus no matter what is the value of $\sigma^2_t$, the highest density mode of true posterior is always the mode $\mu^*$ that is closest to the initial point $x_t$. And the decision boundary is always the vonoroi centered at $\mu^i$.

\textbf{Lemma 3.3}
\textit{The PF-ODE can be written as:}
\begin{gather}
\frac{dX_t}{dt} = \underbrace{\sum_{i=1}^N\frac{w_i}{2}\frac{d\sigma_t^2}{dt} \frac{(X_t-\mu^i)}{\sigma^2+\sigma_t^2}}_{\textrm{velocity field } v_t}, w_i= (\exp{-\frac{||X_t-\mu^i||^2}{2(\sigma^2+\sigma_t^2)}}) / (\sum_{j=1}^N\exp{-\frac{||X_t-\mu^j||^2}{2(\sigma^2+\sigma_t^2)}}) \notag.
\end{gather}
\begin{proof}
    We need to compute the score function first:
    \begin{gather}
        \nabla \log p_{\theta}(X_t=x_t) = \frac{\nabla p_{\theta}(X_t=x_t)}{p_{\theta}(X_t=x_t)} \notag \\
        = \frac{1}{p_{\theta}(X_t=x_t)} \nabla (\sum_{i=1}^N \frac{1}{N}(\mathcal{N}(X_t=x_t|\mu^i,(\sigma^2+\sigma_t^2)I))) \notag \\
        = \frac{1}{p_{\theta}(X_t=x_t)} \sum_{i=1}^N \frac{1}{N}\mathcal{N}(X_t=x_t|\mu^i,(\sigma^2+\sigma_t^2))(-\frac{(x-\mu^i)}{\sigma^2+\sigma_t^2}) \notag \\
        = \sum_{i=1}^N ((\exp{-\frac{||x_t-\mu^i||^2}{2(\sigma^2+\sigma_t^2)}}) / (\sum_{j=1}^N\exp{-\frac{||x_t-\mu^j||^2}{2(\sigma^2+\sigma_t^2)}}))(-\frac{x-\mu^i}{\sigma^2+\sigma_t^2}).
    \end{gather}
    Take the score function into the PF-ODE definition in Eq.~\ref{eq:rsde}, we can obtain the result.
\end{proof}

With those basic properties, we can show that the solution of PF-ODE with initial value $X_t=x_t$ has non-zero density in true posterior $p_{\theta}(X_0|X_t)$.

\textbf{Proposition 3.2} \textit{The solution of PF-ODE has a positive likelihood in true posterior with high probability, \textit{i.e.},
}
\begin{gather}
    p_{\theta}(X_0=\Phi(x_t)|X_t=x_t) \ge \frac{1}{N}\frac{1}{\sqrt{(4\pi\sigma^2)^d}}\exp{(-\frac{2c^2}{\sigma_t^2} - \frac{d+1}{2})} \textrm{, with probability } 1-p.
\end{gather}
\begin{proof}
\begin{gather}
    p_{\theta}(X_0=\Phi(x_t)|X_t=x_t) = \sum_{i=1}^N u_i\mathcal{N}(X_0=\Phi(x_t)|\mu^i,\sigma^2I) \\
    \ge u_j \mathcal{N}(X_0=\Phi(x_t)|\mu^j,\sigma^2 I),\forall j
\end{gather}
We let $k=\min_i \{||\mu^i - \Phi(x_t)||\}$, by assumption we have $||\Phi_0(x_t) - \mu^k|| \le  \sigma^2 + d\sigma^2$ with probability $1-p$.
\begin{gather}
    p_{\theta}(X_0=\Phi(x_t)|X_t=x_t) \ge u_k\frac{1}{\sqrt{(2\pi\sigma^2)^d}}\exp{-\frac{||\mu^k - \Phi(x_t)||^2}{2\sigma^2}} \\
    \overset{(a)}{\ge} u_k\frac{1}{\sqrt{(2\pi\sigma^2)^d}}\exp{-\frac{(d+1)\sigma^2}{2\sigma^2}} \\
    = \frac{\exp{-\frac{||\Phi(x_t)-x_t||^2}{2\sigma_t^2}}}{\sum_{j=1}^N\sqrt{(1+\sigma^2/\sigma_t^2)^d}\exp{-\frac{||x_t-\mu^i||^2}{2(\sigma^2+\sigma_t^2)}}}\frac{1}{\sqrt{(2\pi\sigma^2)^d}}\exp{-\frac{d+1}{2}}\\
    \overset{(b)}{\ge} \frac{\exp{-\frac{4c^2}{2\sigma_t^2}}}{\sqrt{2^d}\sum_{j=1}^N \exp{0}}\frac{1}{\sqrt{(2\pi\sigma^2)^d}}\exp{-\frac{d+1}{2}} \\
    = \frac{1}{N}\frac{1}{\sqrt{(4\pi\sigma^2)^d}}\exp{(-\frac{2c^2}{\sigma_t^2}-\frac{d+1}{2})}
\end{gather}
(a) is due to the assumption that $||\Phi(x_t) - \mu^k||^2\le \sigma^2 + d \sigma^2$. (b) is due to $||\Phi_0(x_t) - x_t|| \le ||\Phi_0(x_t)|| + ||x_t||$. As they are both bounded by $c$, $||\Phi_0(x_t) - x_t||^2$ is bounded by $4c^2$. And $1+\frac{\sigma^2}{\sigma_t^2} \le 2$. 
\end{proof}
We can study the PF-ODE in Eq.~\ref{eq:pfode} informally when $\sigma_t^2$ is rather small or large. When $\sigma_t^2$ is small, the soft-max $w_i$ becomes a "hard"-max. Denote $k=\min_i \{||\mu^i-\Phi(x_t)||\}$ and the PF-ODE at that time can be written as
\begin{gather}
    \frac{dX_t}{dt} = -\frac{1}{2}\frac{d\sigma_t^2}{dt} (-\frac{(X_t-\mu^k)}{\sigma^2+\sigma_t^2}).
\end{gather}
At that time, the PF-ODE is first order separable. And we can solve it with initial value $X_t=x_t$ as 
\begin{gather}
    \frac{\Phi(x_t)-\mu^k}{x_t-\mu^k} = e^{h(t)},
\end{gather}
where $h(.)$ is some function of $t$ related to the $d\sigma_t^2/dt$. 

Let's assume a simple $\sigma_t^2$ schedule such as $\sigma_t^2=t^2$. In that case, we have
\begin{gather}
    \Phi(x_t) = \mu^k + (x_t-\mu^k)e^{\frac{1}{2}\ln \frac{\sigma^2}{\sigma^2 + t^2}}.
\end{gather}
The solution has the form of $\mu^k$ with an offset term weighted by an exponential term. When $\sigma^2$ is small, the exponential term goes to $0$ very fast. And therefore $\Phi(x_t) \approx \mu^k$ at that time. 

\section{Additional Experiment Setup}
\label{app:im}
\subsection{Implementation Details}
All the experiments are implemented in Pytorch, and run in a computer with AMD EPYC 7742 CPU and Nvidia A100 GPU.
\begin{table}[htb]
\centering
\caption{The model specification used for different non-linear operators.}
\begin{tabular}{@{}lcc@{}}
\toprule
                     & Model A & Model B \\ \midrule
Segmentation & MobileNet + C1 & ResNet50 + PPM\\
Layout       & \citet{Lin2018IndoorSL} & \citet{Lin2018IndoorSL} + DA \\
Caption      & BLIP & CLIP \\
Classification   & ResNet50 & VITB16 \\ \bottomrule
\end{tabular}
\label{tab:modelab}
\end{table}

As we have shown, using the same model for $f(.)$ causes overfitting for neural network based $f(.)$. Therefore, we adopt different model for $f(.)$ in DIS and testing, and the details are shown in Tab.~\ref{tab:modelab}.

\begin{table}[htb]
\centering
\caption{Metrics for DIS loss and evaluation.}
\begin{tabular}{@{}lcc@{}}
\toprule
               & $d(.,.)$ for DIS & metric for Test \\ \midrule
Segmentation   & Cross Entropy  & mIOU              \\
Layout         & Cross Entropy  & mIOU \\
Caption        & Cross Entropy  & CLIP score \\
Classification & Cross Entropy  & Accuracy          \\
Downsample     & MSE            & MSE               \\ \bottomrule
\end{tabular}
\end{table}

For different operators, we also have different $d(.,.)$ to evaluate the distance $d(f(x_{0|t}),y)$ during the DIS process. For all four non-linear operators, the cross entropy are used for $d(.,.)$. While for down-sample, we adopt MSE. To evaluate how consistent the generated samples are to $y$, we use y-metrics. Or to say, the metrics computed with input measurement $y$ and $f(x_{0|t})$. More specifically, for Segmentation and Layout, we evaluate consistency by y-mIOU. For image caption, we evaluate consistency by CLIP score \citep{Hessel2021CLIPScoreAR}. For classification, we evaluate consistency by accuracy. And for down-sample, we evaluate consistency by MSE. Note that the $d(.,.)$ used during DIS follows the convention of training corresponding $f(.)$, and the y-metric used for testing also follows the convention of testing corresponding $f(.)$.

\subsection{Details of DIS Algorithm}
Below we provide detailed algorithm of different DIS methods including the ones we compare to and our own.
\begin{minipage}[t]{0.5\textwidth} %
\vspace{0pt}
\IncMargin{1.0em}
\begin{algorithm}[H]
\DontPrintSemicolon
\caption{DPS}\label{alg:dps}
\textbf{procedure} DPS($p_{\theta}(.|.),T,f(.),y,d(.,.),\zeta_t$)\;
$\quad$ $x_T = \mathcal{N}(0,T^2I)$\;
$\quad$\textbf{for} $t=T$ {\bfseries to} $1$ \textbf{do}\;
$\quad\quad$$x_{t-1} \sim p_{\theta}(X_{t-1}|X_t=x_t)$\;
$\quad\quad$$x_{0|t} = \mathbb{E}[X_0|X_t=x_t]$\;
$\quad\quad$$x_{t-1}\leftarrow x_{t-1} - \zeta_t d(f(x_{0|t}), y)$\;
$\quad$\textbf{return} $x_0$\;
\end{algorithm}
\end{minipage}
\begin{minipage}[t]{0.5\textwidth} %
\vspace{0pt}
\IncMargin{1.0em}
\begin{algorithm}[H]
\DontPrintSemicolon
\caption{FreeDOM}\label{alg:fdm}
\textbf{procedure} FreeDOM($p_{\theta}(.|.),q(.|.)$,$T,f(.),y,d(.,.),\zeta_t,r,K$)\;
$\quad$ $x_T = \mathcal{N}(0,T^2I)$\;
$\quad$\textbf{for} $t=T$ {\bfseries to} $1$ \textbf{do}\;
$\quad\quad$\textbf{for} $t'=K$ {\bfseries to} $1$ \textbf{do}\;
$\quad\quad\quad$$x_{t-1} \sim p_{\theta}(X_{t-1}|X_t=x_t)$\;
$\quad\quad\quad$$x_{0|t} = \mathbb{E}[X_0|X_t=x_t]$\;
$\quad\quad\quad$$x_{t-1}\leftarrow x_{t-1} - \zeta_t d(f(x_{0|t}), y)$\;
$\quad\quad\quad$ \textbf{if} $t'\neq 1, t\in r$ \textbf{then}\;
$\quad\quad\quad\quad$ $x_t = q(X_t|X_{t-1} = x_{t-1})$\;
$\quad\quad\quad$ \textbf{else} \;
$\quad\quad\quad\quad$ \textbf{break}\;
$\quad$\textbf{return} $x_0$\;
\end{algorithm}
\end{minipage}
\begin{minipage}[t]{0.5\textwidth} %
\vspace{0pt}
\IncMargin{1.0em}
\begin{algorithm}[H]
\DontPrintSemicolon
\caption{MPGD}\label{alg:mpgd}
\textbf{procedure} MPGD($q_{\theta}(.|.,.),T,f(.),y,d(.,.),\zeta_t$)\;
$\quad$ $x_T = \mathcal{N}(0,T^2I)$\;
$\quad$\textbf{for} $t=T$ {\bfseries to} $1$ \textbf{do}\;
$\quad\quad$$x_{0|t} = \mathbb{E}[X_0|X_t=x_t]$\;
$\quad\quad$$x_{0|t} \leftarrow x_{0|t} - \zeta_t d(f(x_{0|t}), y)$\;
$\quad\quad$$x_{t-1}\leftarrow q(X_{t-1}|X_t=x_t,X_t=x_{0|t})$\;
$\quad$\textbf{return} $x_0$\;
\end{algorithm}
\end{minipage}
\begin{minipage}[t]{0.5\textwidth} %
\vspace{0pt}
\IncMargin{1.0em}
\begin{algorithm}[H]
\DontPrintSemicolon
\caption{LGD (Single Sample)}\label{alg:lgd}
\textbf{procedure} LGD($p_{\theta}(.|.),T,f(.),y,d(.,.),\zeta_t,r_t$)\;
$\quad$ $x_T = \mathcal{N}(0,T^2I)$\;
$\quad$\textbf{for} $t=T$ {\bfseries to} $1$ \textbf{do}\;
$\quad\quad$$x_{t-1} \sim p_{\theta}(X_{t-1}|X_t=x_t)$\;
$\quad\quad$$x_{0|t} = \mathbb{E}[X_0|X_t=x_t] + \mathcal{N}(0,r_t^2I)$\;
$\quad\quad$$x_{t-1}\leftarrow x_{t-1} - \zeta_t d(f(x_{0|t}), y)$\;
$\quad$\textbf{return} $x_0$\;
\end{algorithm}
\end{minipage}
\begin{minipage}[t]{0.5\textwidth} %
\vspace{0pt}
\IncMargin{1.0em}
\begin{algorithm}[H]
\DontPrintSemicolon
\caption{STSL (Single Sample)}\label{alg:stsl}
\textbf{procedure} STSL($p_{\theta}(.|.),T,f(.),y,d(.,.),\zeta_t,\eta_t$)\;
$\quad$ $x_T = \mathcal{N}(0,T^2I)$\;
$\quad$\textbf{for} $t=T$ {\bfseries to} $1$ \textbf{do}\;
$\quad\quad$$x_{0|t} = \mathbb{E}[X_0|X_t=x_t]$\;
$\quad\quad$$x_{t}\leftarrow x_{t} - \zeta_t d(f(x_{0|t}), y)$\;
$\quad\quad$$\epsilon \sim \mathcal{N}(0,I)$\;
$\quad\quad$$x_{t}\leftarrow x_{t} - \eta_t \nabla_{x_t}(\epsilon^T(s_{\theta}(t,x_t + \epsilon) - s_{\theta}(t,x_t)))$\;
$\quad\quad$$x_{t-1}\sim p_{\theta}(X_{t-1}|X_t=x_t)$\;
$\quad$\textbf{return} $x_0$\;
\end{algorithm}
\end{minipage}
\begin{minipage}[t]{0.5\textwidth} %
\vspace{0pt}
\IncMargin{1.0em}
\begin{algorithm}[H]
\DontPrintSemicolon
\caption{Proposed I}\label{alg:dpscm}
\textbf{procedure} Proposed-I($p_{\theta}(.|.),T,f(.),y,d(.,.),\zeta_t,g_{\theta}(.,.),\tau$)\;
$\quad$ $x_T = \mathcal{N}(0,T^2I)$\;
$\quad$\textbf{for} $t=T$ {\bfseries to} $1$ \textbf{do}\;
$\quad\quad$$x_{t-1} \sim p_{\theta}(X_t|X_{t-1}=x_{t-1})$\;
$\quad\quad$$x_{0|t} = g_{\theta}(t,x_t)+\mathcal{N}(0,\tau^2I)$\;
$\quad\quad$$x_{t-1}\leftarrow x_{t-1} - \zeta_t d(f(x_{0|t}), y)$\;
$\quad$\textbf{return} $x_0$\;
\end{algorithm}
\end{minipage}
\begin{minipage}[t]{0.5\textwidth}
\vspace{0pt}
\IncMargin{1.0em}
\begin{algorithm}[H]
\DontPrintSemicolon
\caption{Consistency Model}\label{alg:cm}
\textbf{procedure} CM($g_\theta(.,.),t_{1:N}$)\;
$\quad$\textbf{for} $n=1$ {\bfseries to} $N-1$ \textbf{do}\;
$\quad\quad$$z\sim \mathcal{N}(0,I)$\;
$\quad\quad$$x_{t_n} \leftarrow x_0 + t_n z$\;
$\quad\quad$$x_0\leftarrow g_\theta(t_n,x_{t_n})$\;
$\quad$\textbf{return} $x_0$\;
\end{algorithm}
\end{minipage}
\begin{minipage}[t]{0.5\textwidth}
\vspace{0pt}
\IncMargin{1.0em}
\begin{algorithm}[H]
\DontPrintSemicolon
\caption{Proposed II}\label{alg:cmi}
\textbf{procedure} CMInversion($g_{\theta}(.,.)$, $t_{1:N}$, $f(.)$, $y$, $\tau$)\;
$\quad$\textbf{for} $n=1$ {\bfseries to} $N-1$ \textbf{do}\;
$\quad\quad$$z\sim \mathcal{N}(0,I)$\;
$\quad\quad$\textbf{for} $k=1$ {\bfseries to} K \textbf{do}\;
$\quad\quad\quad$ $x_{t_n} \leftarrow x_0 + t_n z$\;
$\quad\quad\quad$ $x_0\leftarrow g_\theta(t_n,x_{t_n})$\;
$\quad\quad\quad$ $x_{\tau} \leftarrow x_0 + \mathcal{N}(0,\tau^2I)$\;
$\quad\quad\quad$ $z \leftarrow z - \zeta \frac{d}{d z}d(f(x_{\tau}),y)$\;
$\quad$ {\bfseries return} $x_0$
\end{algorithm}
\end{minipage}

\textbf{FreeDOM} \citet{Yu2023FreeDoMTE} propose to adopt the time-travel that is designed specifically for in-painting \citep{Lugmayr2022RePaintIU} to general operator $f(.)$. (See Algorithm.~\ref{alg:fdm}) More specifically, it proposes an inner loop that goes forward after a backward step with forward kernel $q(.|.)$. The new hyper-parameters are time-travel steps $K$ and time-travel range $r$.

\textbf{MPGD} \citet{He2023ManifoldPG} propose to perform the gradient ascent directly on posterior mean instead of on $x_t$. And the posterior mean after gradient ascent is used to correct the score function (See Algorithm.~\ref{alg:mpgd}). They claim that their approach is able to converge faster and outperform DPS when $T=20,100$. However, as we use $T=1000$, the advantage of their approach is not clearly shown in our experiments.

\textbf{LGD} \citet{Song2023LossGuidedDM} adopt a Gaussian approximation to posterior sample (See Algorithm.~\ref{alg:lgd}). More specifically, they use an additive of Gaussian noise and posterior mean as an approximation of Gaussian sample. And the mean of approximated sample is the same as real posterior sample. This approach is later improved by \citet{boys2023tweedie,Rout2023BeyondFT} to second order. Or to say, they estimate the covariance of Gaussian using second order Tweedie's formula. And the mean and covariance of approximated sample is the same as real posterior sample. The authors of LGD further propose a multi-sample approach to reduce gradient variance. However, we only use LGD with sample size $n=1$ for fair comparison. The new hyper-parameters is variance $\tau$.

\textbf{STSL} \citet{Rout2023BeyondFT} propose to improve \citet{Song2023LossGuidedDM} by estimating the posterior as Gaussian distribution with posterior mean and posterior covariance. As directly estimating the posterior covariance using second order Tweedie's formula is expensive, they propose a Monte Carlo estimation and the resulting algorithm is shown in Algorithm.~\ref{alg:stsl}. However, we only use STSL with sample size $n=1$ for fair comparison. 
\subsection{Hyper-parameters}

\begin{table*}[thb]
\centering
\resizebox{\linewidth}{!}{
\begin{tabular}{@{}lccccc@{}}
\toprule
                             & Segmentation & Layout & Caption & Classification & Down-sampling \\ \midrule
DPS & $\zeta=256.0$ & $\zeta=7.2$ & $\zeta=24.0$ & $\zeta=8.0$ & $\zeta=14.4$ \\ \midrule
\multirow{2}{*}{LGD} & $\zeta=256.0$ & $\zeta=7.2$ & $\zeta=24.0$ & $\zeta=8.0$ & $\zeta=14.4$ \\
                             & \multicolumn{5}{c}{$K=1,\tau=0.2$} \\ \midrule
\multirow{2}{*}{FreeDOM} & $\zeta=256.0$ & $\zeta=7.2$ & $\zeta=24.0$ & $\zeta=8.0$ & $\zeta=14.4$ \\
                             & \multicolumn{5}{c}{$K=2,r=[100,200]$}                                          \\ \midrule
MPGD                         & $\zeta=2560.0$ & $\zeta=72.0$ & $\zeta=240.0$ & $\zeta=80.0$ & $\zeta=144.0$ \\ \midrule
\multirow{2}{*}{Proposed I}  & $\zeta=256.0$ & $\zeta=7.2$ & $\zeta=24.0$ & $\zeta=8.0$ & $\zeta=14.4$ \\ 
                             & \multicolumn{5}{c}{$\tau=0.2$}                                          \\ \midrule
\multirow{3}{*}{Proposed II} & $\zeta_1=0.1,\zeta_2=0.005$ & $\zeta_1=0.1,\ \zeta_2=0.001$ & $\zeta_1=0.1,\ \zeta_2=0.001$ & $\zeta_1=0.1,\zeta_2=0.005$       & $\zeta_1=0.3,\zeta_2=0.03$ \\
                             & $ts=3,6,...,18,30,39$ & $ts=75,100,125,150$ & $ts=75,100,125,150$ & $ts=25,50,75,125,150$ & $ts=75,100,125,150$ \\
                             & $K=40$ & $K=151$ & $K=151$ &  $K=151$    & $K=151$ \\ \bottomrule
\end{tabular}
}
\caption{The hyper-parameters of other DIS and proposed approaches.}
\label{tab:hp}
\end{table*}
\begin{figure}[thb]
\centering
    \includegraphics[width=0.6\linewidth]{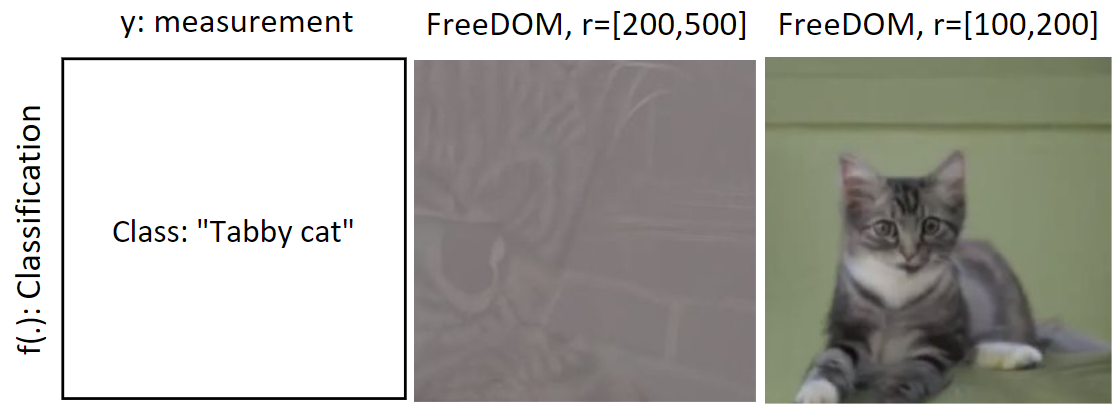}
\caption{Comparison of different inner-loop range for FreeDOM.}
\label{fig:fdm}
\end{figure}
We list the detailed hyper-parameters of DPS, LGD, FreeDOM, MPGD \citep{Chung2022DiffusionPS,Song2023LossGuidedDM,Yu2023FreeDoMTE,he2023manifold} and two of our proposed approaches in Tab.~\ref{tab:hp}. For all the methods, one common hyper-parameter is the step size $\zeta$ used in gradient descent. For LGD and our Proposed I, an additional hyper-parameter is the additional additive noise $\tau=0.2$. We do not use the multi-sample LGD as it is significantly slower than all other approaches. For FreeDOM, the additional parameters are time-travel steps $K$, and time-travel range $r$. We set $K=2$ for fair comparison, as a large $K$ make FreeDOM significantly slower than all other approaches. We set $r=[100,200]$ instead of $r=[200,500]$ in original paper \citep{Yu2023FreeDoMTE}. This is because we find that setting $r=[200,500]$ in VE-diffusion has significant negative effect on sample quality. The visual comparison is in Fig.~\ref{fig:fdm}. For Proposed II, in addition to learning rate $\zeta$, we also control the timestep schedule $ts$ and optimization step $K$. 

\section{Additional Experiment Results}
\subsection{Additional Visual Results}
\begin{figure*}[htb]
\centering
    \includegraphics[width=\linewidth]{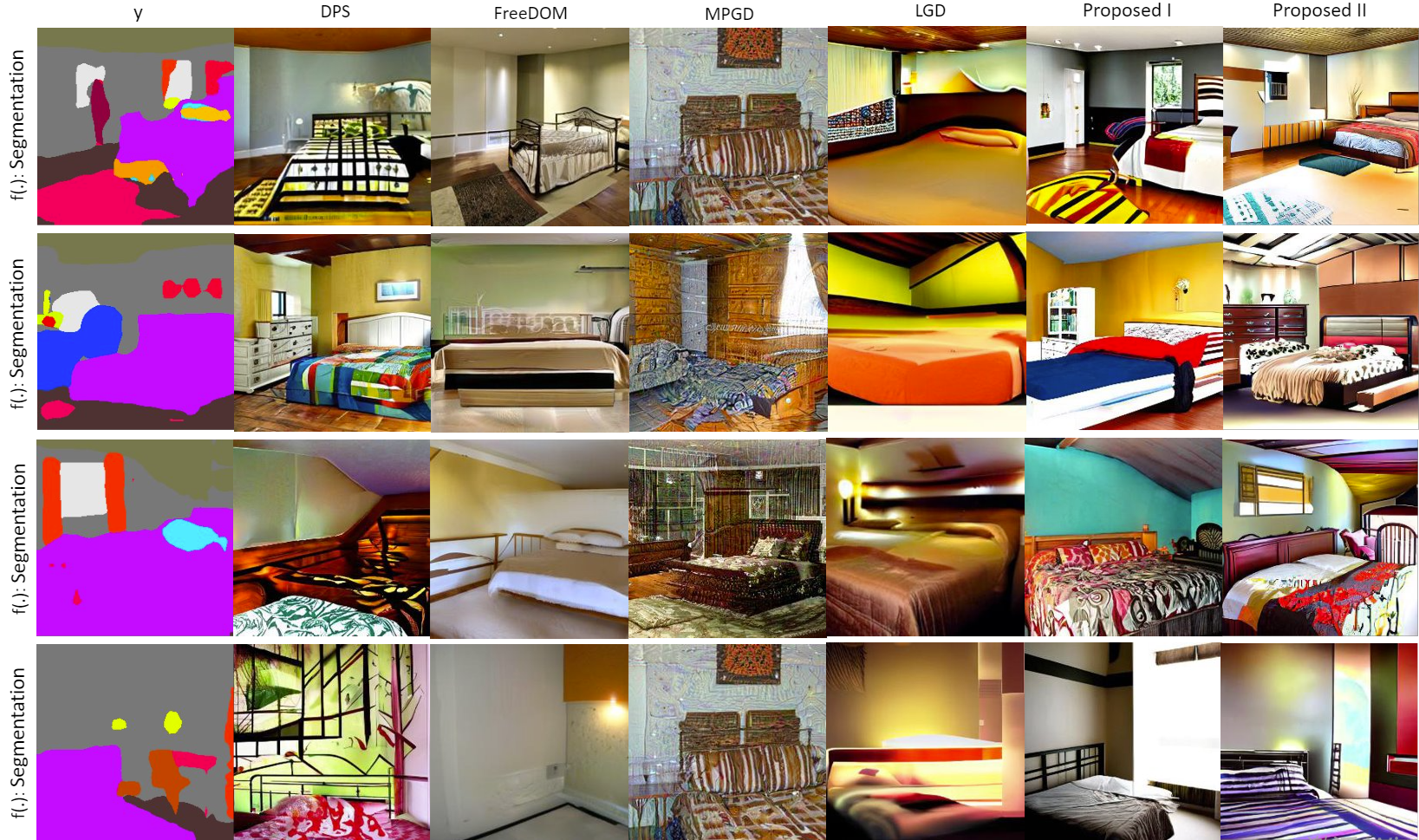}
    \includegraphics[width=\linewidth]{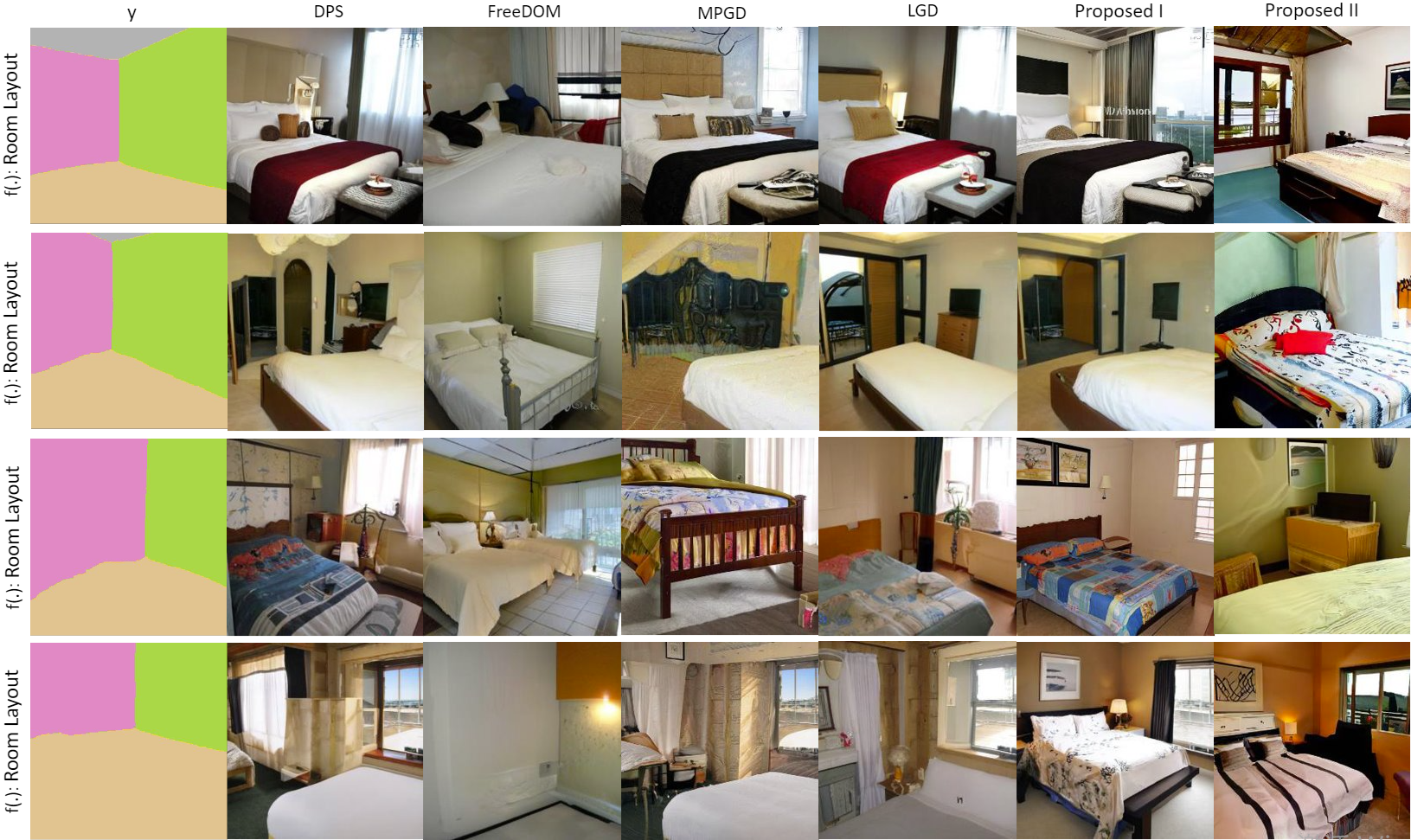}
\caption{Additional visual results on image segmentation and layout estimation.}
\label{fig:avr1}
\end{figure*}

\begin{figure*}[htb]
\centering
    \includegraphics[width=\linewidth]{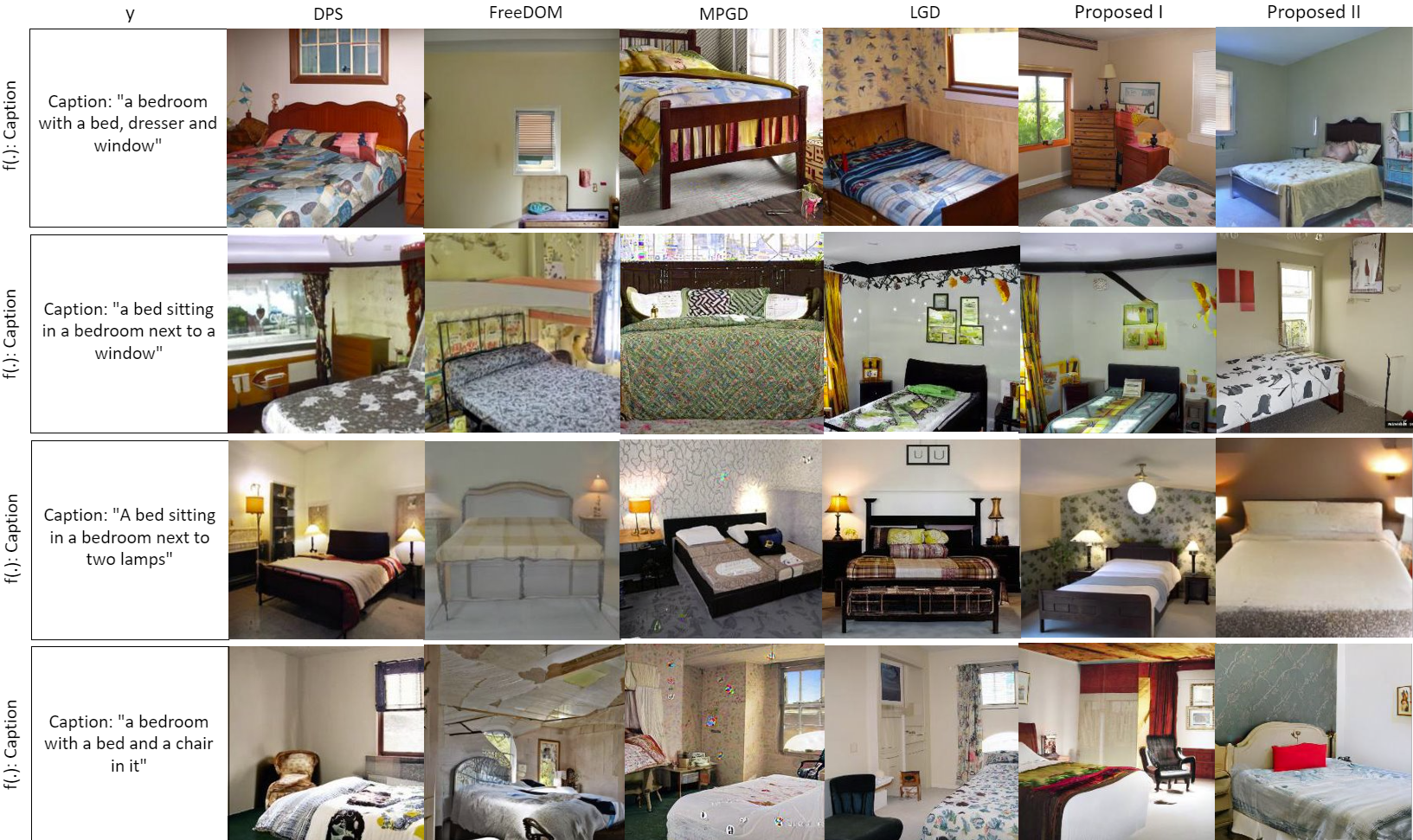}
    \includegraphics[width=\linewidth]{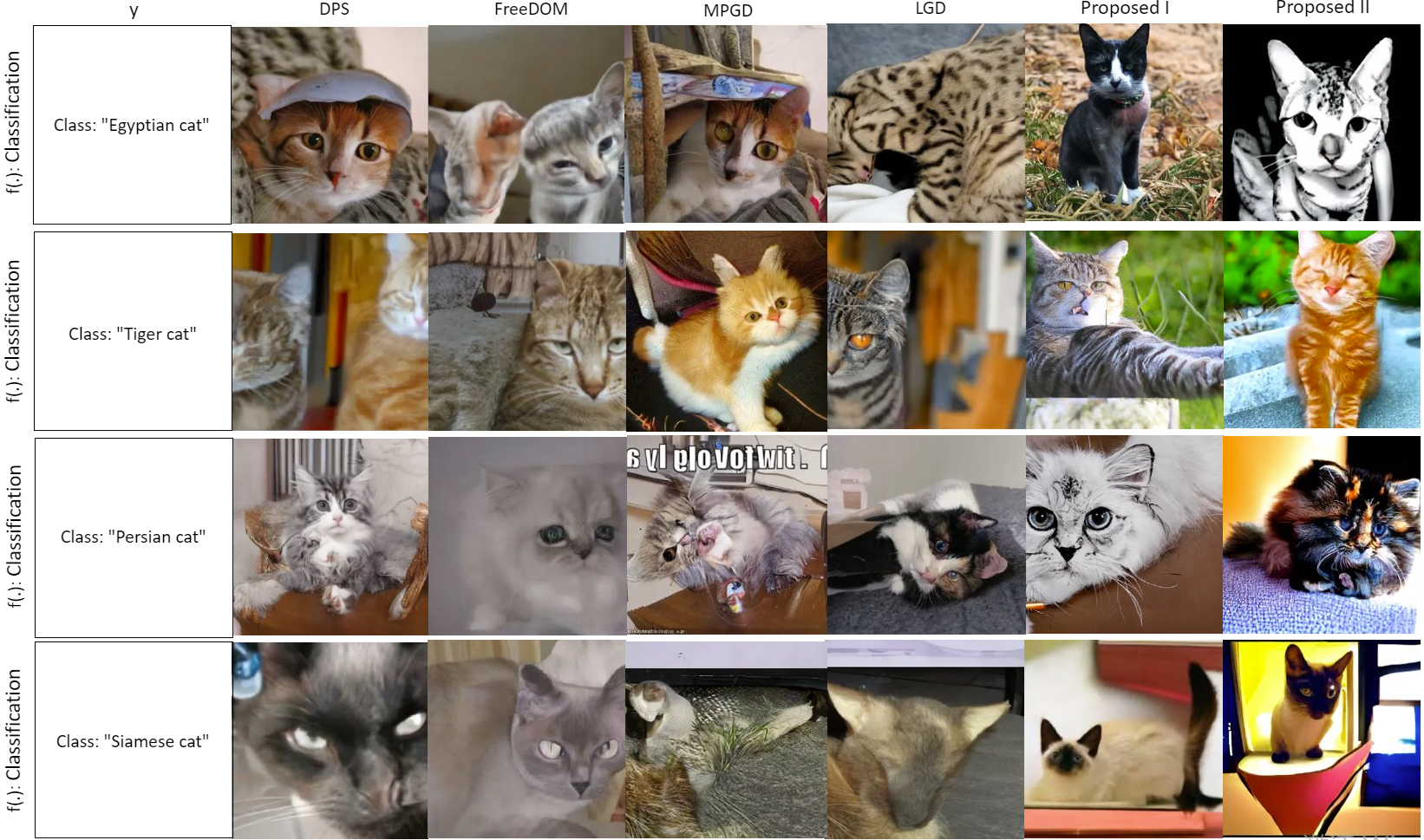}
\caption{Additional visual results on image captioning and image classification.}
\label{fig:avr3}
\end{figure*}

We present more visual results of non-linear operators in Fig.~\ref{fig:avr1} and \ref{fig:avr3}. It is shown that our Proposed I has best consistency with measurement $y$ and best sample quality for most of the times. Our Proposed II also has a good consistency and sample quality.

\begin{figure}[htb]
\centering
    \includegraphics[width=0.8\linewidth]{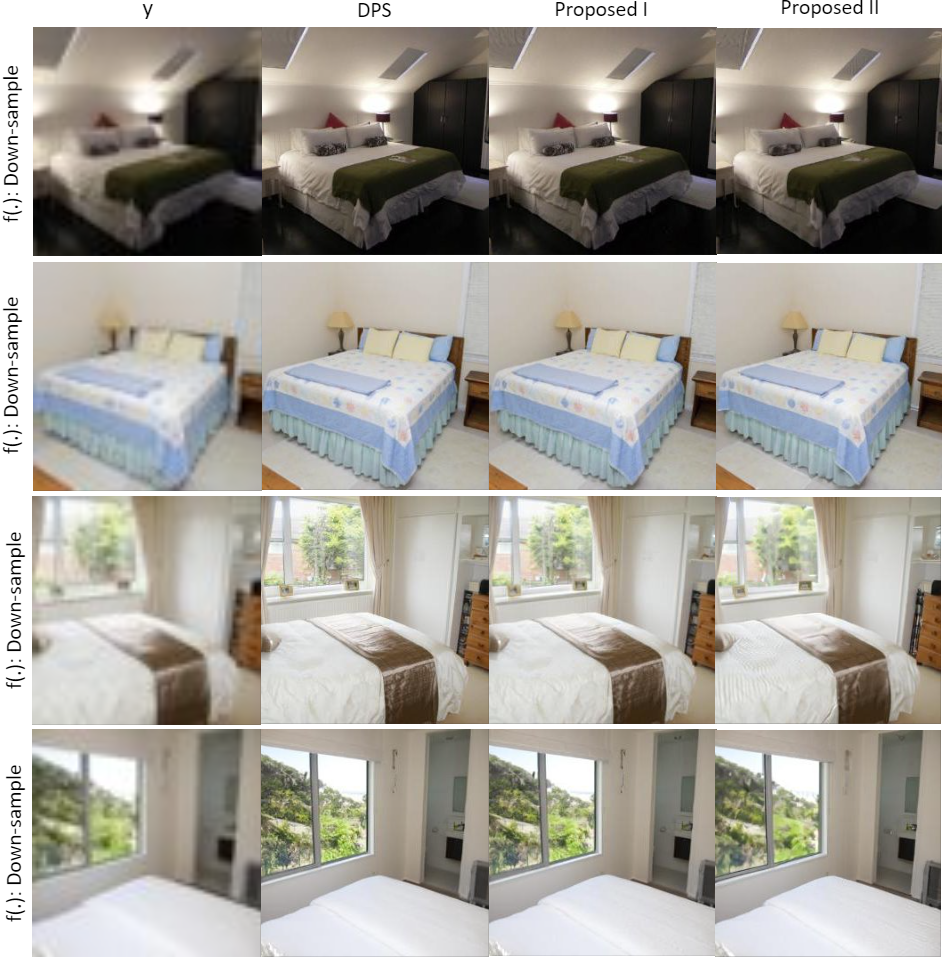}
\caption{Additional visual results on image down-sampling.}
\label{fig:avrsr}
\end{figure}

Furthermore, we present more visual results of down-sampling in Fig.~\ref{fig:avrsr}. It is shown that both our Proposed II and Proposed II work as good as DPS.

\subsection{Failure Cases}
\begin{figure}[htb]
\centering
    \includegraphics[width=0.6\linewidth]{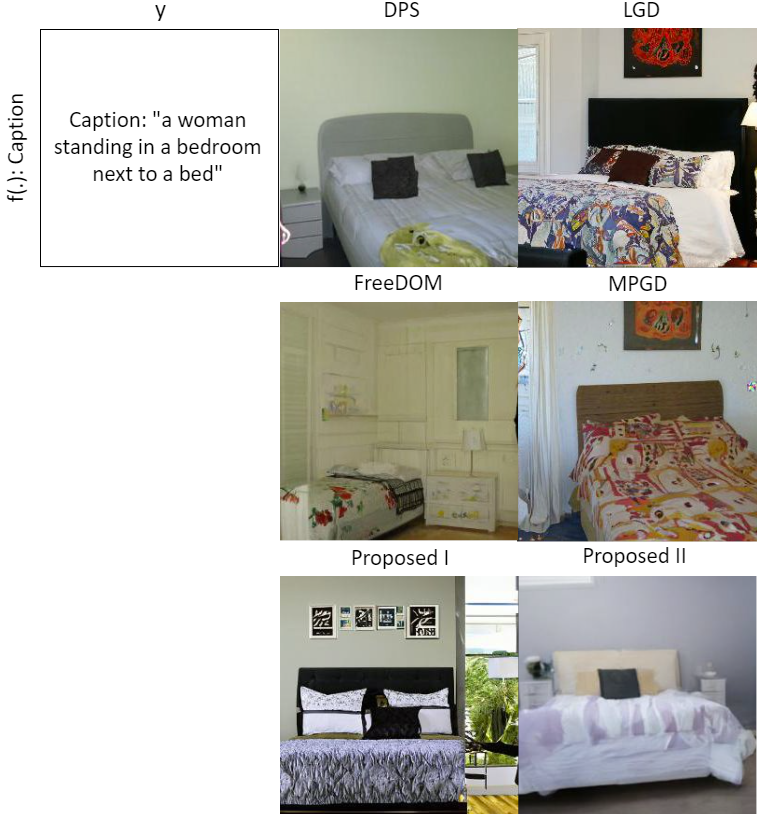}
\caption{Visual results of a failure case.}
\label{fig:fc}
\end{figure}

When the measurement $y$ is too far from diffusion prior, our approaches and other DIS approaches fail. An example of such failure is shown in Fig.~\ref{fig:fc}. The input measurement $y$ describes a woman. However, human is not a part of LSUN bedroom dataset. And none-of the DIS approaches is able to generate a woman. And the samples generated by DIS look like unconditional sample.

\section{Additional Discussion}

\subsection{Reproducibility Statement}
The proof of all theoretical results are shown in Appendix.~\ref{app:pf}. For experiments, all two datasets are publicly available. In Appendix.~\ref{app:im}, we provide additional implementation details of all other DIS that we compare to. Further, detailed hyper-parameters of all baselines and our proposed approach are presented. Besides, we provide source code for reproducing empirical results as supplementary material.

\subsection{Broader Impact}
The approach proposed in this paper allows conditional generation without training a new model. This saves the energy of training conditional generative diffusion model and reduces the carbon emission. Potential negative impact is the same as other conditional generative model, such as trustworthiness brought by generating fake image.

\clearpage

\newpage
\section*{NeurIPS Paper Checklist}

\begin{enumerate}

\item {\bf Claims}
    \item[] Question: Do the main claims made in the abstract and introduction accurately reflect the paper's contributions and scope?
    \item[] Answer: \answerYes{} 
    \item[] Justification: Yes.
    \item[] Guidelines:
    \begin{itemize}
        \item The answer NA means that the abstract and introduction do not include the claims made in the paper.
        \item The abstract and/or introduction should clearly state the claims made, including the contributions made in the paper and important assumptions and limitations. A No or NA answer to this question will not be perceived well by the reviewers. 
        \item The claims made should match theoretical and experimental results, and reflect how much the results can be expected to generalize to other settings. 
        \item It is fine to include aspirational goals as motivation as long as it is clear that these goals are not attained by the paper. 
    \end{itemize}

\item {\bf Limitations}
    \item[] Question: Does the paper discuss the limitations of the work performed by the authors?
    \item[] Answer: \answerYes{} 
    \item[] Justification: In Discussion.
    \item[] Guidelines:
    \begin{itemize}
        \item The answer NA means that the paper has no limitation while the answer No means that the paper has limitations, but those are not discussed in the paper. 
        \item The authors are encouraged to create a separate "Limitations" section in their paper.
        \item The paper should point out any strong assumptions and how robust the results are to violations of these assumptions (e.g., independence assumptions, noiseless settings, model well-specification, asymptotic approximations only holding locally). The authors should reflect on how these assumptions might be violated in practice and what the implications would be.
        \item The authors should reflect on the scope of the claims made, e.g., if the approach was only tested on a few datasets or with a few runs. In general, empirical results often depend on implicit assumptions, which should be articulated.
        \item The authors should reflect on the factors that influence the performance of the approach. For example, a facial recognition algorithm may perform poorly when image resolution is low or images are taken in low lighting. Or a speech-to-text system might not be used reliably to provide closed captions for online lectures because it fails to handle technical jargon.
        \item The authors should discuss the computational efficiency of the proposed algorithms and how they scale with dataset size.
        \item If applicable, the authors should discuss possible limitations of their approach to address problems of privacy and fairness.
        \item While the authors might fear that complete honesty about limitations might be used by reviewers as grounds for rejection, a worse outcome might be that reviewers discover limitations that aren't acknowledged in the paper. The authors should use their best judgment and recognize that individual actions in favor of transparency play an important role in developing norms that preserve the integrity of the community. Reviewers will be specifically instructed to not penalize honesty concerning limitations.
    \end{itemize}

\item {\bf Theory Assumptions and Proofs}
    \item[] Question: For each theoretical result, does the paper provide the full set of assumptions and a complete (and correct) proof?
    \item[] Answer: \answerYes{} 
    \item[] Justification: In appendix.
    \item[] Guidelines:
    \begin{itemize}
        \item The answer NA means that the paper does not include theoretical results. 
        \item All the theorems, formulas, and proofs in the paper should be numbered and cross-referenced.
        \item All assumptions should be clearly stated or referenced in the statement of any theorems.
        \item The proofs can either appear in the main paper or the supplemental material, but if they appear in the supplemental material, the authors are encouraged to provide a short proof sketch to provide intuition. 
        \item Inversely, any informal proof provided in the core of the paper should be complemented by formal proofs provided in appendix or supplemental material.
        \item Theorems and Lemmas that the proof relies upon should be properly referenced. 
    \end{itemize}

    \item {\bf Experimental Result Reproducibility}
    \item[] Question: Does the paper fully disclose all the information needed to reproduce the main experimental results of the paper to the extent that it affects the main claims and/or conclusions of the paper (regardless of whether the code and data are provided or not)?
    \item[] Answer: \answerYes{} 
    \item[] Justification: Code is available in supplementary materials.
    \item[] Guidelines:
    \begin{itemize}
        \item The answer NA means that the paper does not include experiments.
        \item If the paper includes experiments, a No answer to this question will not be perceived well by the reviewers: Making the paper reproducible is important, regardless of whether the code and data are provided or not.
        \item If the contribution is a dataset and/or model, the authors should describe the steps taken to make their results reproducible or verifiable. 
        \item Depending on the contribution, reproducibility can be accomplished in various ways. For example, if the contribution is a novel architecture, describing the architecture fully might suffice, or if the contribution is a specific model and empirical evaluation, it may be necessary to either make it possible for others to replicate the model with the same dataset, or provide access to the model. In general. releasing code and data is often one good way to accomplish this, but reproducibility can also be provided via detailed instructions for how to replicate the results, access to a hosted model (e.g., in the case of a large language model), releasing of a model checkpoint, or other means that are appropriate to the research performed.
        \item While NeurIPS does not require releasing code, the conference does require all submissions to provide some reasonable avenue for reproducibility, which may depend on the nature of the contribution. For example
        \begin{enumerate}
            \item If the contribution is primarily a new algorithm, the paper should make it clear how to reproduce that algorithm.
            \item If the contribution is primarily a new model architecture, the paper should describe the architecture clearly and fully.
            \item If the contribution is a new model (e.g., a large language model), then there should either be a way to access this model for reproducing the results or a way to reproduce the model (e.g., with an open-source dataset or instructions for how to construct the dataset).
            \item We recognize that reproducibility may be tricky in some cases, in which case authors are welcome to describe the particular way they provide for reproducibility. In the case of closed-source models, it may be that access to the model is limited in some way (e.g., to registered users), but it should be possible for other researchers to have some path to reproducing or verifying the results.
        \end{enumerate}
    \end{itemize}

\item {\bf Open access to data and code}
    \item[] Question: Does the paper provide open access to the data and code, with sufficient instructions to faithfully reproduce the main experimental results, as described in supplemental material?
    \item[] Answer: \answerYes{} 
    \item[] Justification: In supplementary materials.
    \item[] Guidelines:
    \begin{itemize}
        \item The answer NA means that paper does not include experiments requiring code.
        \item Please see the NeurIPS code and data submission guidelines (\url{https://nips.cc/public/guides/CodeSubmissionPolicy}) for more details.
        \item While we encourage the release of code and data, we understand that this might not be possible, so “No” is an acceptable answer. Papers cannot be rejected simply for not including code, unless this is central to the contribution (e.g., for a new open-source benchmark).
        \item The instructions should contain the exact command and environment needed to run to reproduce the results. See the NeurIPS code and data submission guidelines (\url{https://nips.cc/public/guides/CodeSubmissionPolicy}) for more details.
        \item The authors should provide instructions on data access and preparation, including how to access the raw data, preprocessed data, intermediate data, and generated data, etc.
        \item The authors should provide scripts to reproduce all experimental results for the new proposed method and baselines. If only a subset of experiments are reproducible, they should state which ones are omitted from the script and why.
        \item At submission time, to preserve anonymity, the authors should release anonymized versions (if applicable).
        \item Providing as much information as possible in supplemental material (appended to the paper) is recommended, but including URLs to data and code is permitted.
    \end{itemize}

\item {\bf Experimental Setting/Details}
    \item[] Question: Does the paper specify all the training and test details (e.g., data splits, hyperparameters, how they were chosen, type of optimizer, etc.) necessary to understand the results?
    \item[] Answer: \answerYes{} 
    \item[] Justification: In Experimental setup and appendix.
    \item[] Guidelines:
    \begin{itemize}
        \item The answer NA means that the paper does not include experiments.
        \item The experimental setting should be presented in the core of the paper to a level of detail that is necessary to appreciate the results and make sense of them.
        \item The full details can be provided either with the code, in appendix, or as supplemental material.
    \end{itemize}

\item {\bf Experiment Statistical Significance}
    \item[] Question: Does the paper report error bars suitably and correctly defined or other appropriate information about the statistical significance of the experiments?
    \item[] Answer: \answerNo{} 
    \item[] Justification: Error bar is expensive and rarely reported in our community.
    \item[] Guidelines:
    \begin{itemize}
        \item The answer NA means that the paper does not include experiments.
        \item The authors should answer "Yes" if the results are accompanied by error bars, confidence intervals, or statistical significance tests, at least for the experiments that support the main claims of the paper.
        \item The factors of variability that the error bars are capturing should be clearly stated (for example, train/test split, initialization, random drawing of some parameter, or overall run with given experimental conditions).
        \item The method for calculating the error bars should be explained (closed form formula, call to a library function, bootstrap, etc.)
        \item The assumptions made should be given (e.g., Normally distributed errors).
        \item It should be clear whether the error bar is the standard deviation or the standard error of the mean.
        \item It is OK to report 1-sigma error bars, but one should state it. The authors should preferably report a 2-sigma error bar than state that they have a 96\% CI, if the hypothesis of Normality of errors is not verified.
        \item For asymmetric distributions, the authors should be careful not to show in tables or figures symmetric error bars that would yield results that are out of range (e.g. negative error rates).
        \item If error bars are reported in tables or plots, The authors should explain in the text how they were calculated and reference the corresponding figures or tables in the text.
    \end{itemize}

\item {\bf Experiments Compute Resources}
    \item[] Question: For each experiment, does the paper provide sufficient information on the computer resources (type of compute workers, memory, time of execution) needed to reproduce the experiments?
    \item[] Answer: \answerYes{} 
    \item[] Justification: In Experimental setup and appendix.
    \item[] Guidelines:
    \begin{itemize}
        \item The answer NA means that the paper does not include experiments.
        \item The paper should indicate the type of compute workers CPU or GPU, internal cluster, or cloud provider, including relevant memory and storage.
        \item The paper should provide the amount of compute required for each of the individual experimental runs as well as estimate the total compute. 
        \item The paper should disclose whether the full research project required more compute than the experiments reported in the paper (e.g., preliminary or failed experiments that didn't make it into the paper). 
    \end{itemize}
    
\item {\bf Code Of Ethics}
    \item[] Question: Does the research conducted in the paper conform, in every respect, with the NeurIPS Code of Ethics \url{https://neurips.cc/public/EthicsGuidelines}?
    \item[] Answer: \answerYes{} 
    \item[] Justification: Yes.
    \item[] Guidelines:
    \begin{itemize}
        \item The answer NA means that the authors have not reviewed the NeurIPS Code of Ethics.
        \item If the authors answer No, they should explain the special circumstances that require a deviation from the Code of Ethics.
        \item The authors should make sure to preserve anonymity (e.g., if there is a special consideration due to laws or regulations in their jurisdiction).
    \end{itemize}

\item {\bf Broader Impacts}
    \item[] Question: Does the paper discuss both potential positive societal impacts and negative societal impacts of the work performed?
    \item[] Answer: \answerYes{} 
    \item[] Justification: In appendix.
    \item[] Guidelines:
    \begin{itemize}
        \item The answer NA means that there is no societal impact of the work performed.
        \item If the authors answer NA or No, they should explain why their work has no societal impact or why the paper does not address societal impact.
        \item Examples of negative societal impacts include potential malicious or unintended uses (e.g., disinformation, generating fake profiles, surveillance), fairness considerations (e.g., deployment of technologies that could make decisions that unfairly impact specific groups), privacy considerations, and security considerations.
        \item The conference expects that many papers will be foundational research and not tied to particular applications, let alone deployments. However, if there is a direct path to any negative applications, the authors should point it out. For example, it is legitimate to point out that an improvement in the quality of generative models could be used to generate deepfakes for disinformation. On the other hand, it is not needed to point out that a generic algorithm for optimizing neural networks could enable people to train models that generate Deepfakes faster.
        \item The authors should consider possible harms that could arise when the technology is being used as intended and functioning correctly, harms that could arise when the technology is being used as intended but gives incorrect results, and harms following from (intentional or unintentional) misuse of the technology.
        \item If there are negative societal impacts, the authors could also discuss possible mitigation strategies (e.g., gated release of models, providing defenses in addition to attacks, mechanisms for monitoring misuse, mechanisms to monitor how a system learns from feedback over time, improving the efficiency and accessibility of ML).
    \end{itemize}
    
\item {\bf Safeguards}
    \item[] Question: Does the paper describe safeguards that have been put in place for responsible release of data or models that have a high risk for misuse (e.g., pretrained language models, image generators, or scraped datasets)?
    \item[] Answer: \answerNA{} 
    \item[] Justification: No new model is trained.
    \item[] Guidelines:
    \begin{itemize}
        \item The answer NA means that the paper poses no such risks.
        \item Released models that have a high risk for misuse or dual-use should be released with necessary safeguards to allow for controlled use of the model, for example by requiring that users adhere to usage guidelines or restrictions to access the model or implementing safety filters. 
        \item Datasets that have been scraped from the Internet could pose safety risks. The authors should describe how they avoided releasing unsafe images.
        \item We recognize that providing effective safeguards is challenging, and many papers do not require this, but we encourage authors to take this into account and make a best faith effort.
    \end{itemize}

\item {\bf Licenses for existing assets}
    \item[] Question: Are the creators or original owners of assets (e.g., code, data, models), used in the paper, properly credited and are the license and terms of use explicitly mentioned and properly respected?
    \item[] Answer: \answerYes{} 
    \item[] Justification: In Experimental setup and appendix.
    \item[] Guidelines:
    \begin{itemize}
        \item The answer NA means that the paper does not use existing assets.
        \item The authors should cite the original paper that produced the code package or dataset.
        \item The authors should state which version of the asset is used and, if possible, include a URL.
        \item The name of the license (e.g., CC-BY 4.0) should be included for each asset.
        \item For scraped data from a particular source (e.g., website), the copyright and terms of service of that source should be provided.
        \item If assets are released, the license, copyright information, and terms of use in the package should be provided. For popular datasets, \url{paperswithcode.com/datasets} has curated licenses for some datasets. Their licensing guide can help determine the license of a dataset.
        \item For existing datasets that are re-packaged, both the original license and the license of the derived asset (if it has changed) should be provided.
        \item If this information is not available online, the authors are encouraged to reach out to the asset's creators.
    \end{itemize}

\item {\bf New Assets}
    \item[] Question: Are new assets introduced in the paper well documented and is the documentation provided alongside the assets?
    \item[] Answer: \answerYes{} 
    \item[] Justification: In supplementary material
    \item[] Guidelines:
    \begin{itemize}
        \item The answer NA means that the paper does not release new assets.
        \item Researchers should communicate the details of the dataset/code/model as part of their submissions via structured templates. This includes details about training, license, limitations, etc. 
        \item The paper should discuss whether and how consent was obtained from people whose asset is used.
        \item At submission time, remember to anonymize your assets (if applicable). You can either create an anonymized URL or include an anonymized zip file.
    \end{itemize}

\item {\bf Crowdsourcing and Research with Human Subjects}
    \item[] Question: For crowdsourcing experiments and research with human subjects, does the paper include the full text of instructions given to participants and screenshots, if applicable, as well as details about compensation (if any)? 
    \item[] Answer: \answerNA{} 
    \item[] Justification: \answerNA{}
    \item[] Guidelines:
    \begin{itemize}
        \item The answer NA means that the paper does not involve crowdsourcing nor research with human subjects.
        \item Including this information in the supplemental material is fine, but if the main contribution of the paper involves human subjects, then as much detail as possible should be included in the main paper. 
        \item According to the NeurIPS Code of Ethics, workers involved in data collection, curation, or other labor should be paid at least the minimum wage in the country of the data collector. 
    \end{itemize}

\item {\bf Institutional Review Board (IRB) Approvals or Equivalent for Research with Human Subjects}
    \item[] Question: Does the paper describe potential risks incurred by study participants, whether such risks were disclosed to the subjects, and whether Institutional Review Board (IRB) approvals (or an equivalent approval/review based on the requirements of your country or institution) were obtained?
    \item[] Answer: \answerNA{} 
    \item[] Justification: \answerNA{}
    \item[] Guidelines:
    \begin{itemize}
        \item The answer NA means that the paper does not involve crowdsourcing nor research with human subjects.
        \item Depending on the country in which research is conducted, IRB approval (or equivalent) may be required for any human subjects research. If you obtained IRB approval, you should clearly state this in the paper. 
        \item We recognize that the procedures for this may vary significantly between institutions and locations, and we expect authors to adhere to the NeurIPS Code of Ethics and the guidelines for their institution. 
        \item For initial submissions, do not include any information that would break anonymity (if applicable), such as the institution conducting the review.
    \end{itemize}

\end{enumerate}

\end{document}